\documentclass[5p,twocolumn,12pt,authoryear]{elsarticle}

\usepackage{xcolor,soul,framed}

\colorlet{shadecolor}{yellow}

\usepackage[cmex10]{amsmath}
\usepackage{array}
\usepackage{graphicx}
\usepackage{mdwmath}
\usepackage{mdwtab}
\usepackage{eqparbox}
\usepackage{url}
\usepackage{booktabs}
\usepackage{tabularx}
\usepackage{float}
\usepackage{siunitx}
\usepackage{caption}
\usepackage{multirow}
\usepackage{pifont}
\usepackage{makecell}
\usepackage{IEEEtrantools}
\usepackage{ragged2e}
\usepackage{amssymb}
\usepackage{hyperref}
\usepackage{stfloats}
\usepackage{algorithm}
\usepackage{algpseudocode}
\usepackage{placeins}
\usepackage{rotating}

\newcolumntype{L}[1]{>{\RaggedRight\arraybackslash}p{#1}}
\newcolumntype{Y}{>{\RaggedRight\arraybackslash}X}

\newcommand{\cmark}{\ding{51}}
\newcommand{\xmark}{\ding{55}}
\renewcommand{\arraystretch}{1.18}
\begin{document}
\begin{frontmatter}
    \title{Perturbation-Aware Diffusion-Guided Hybrid Segmentation for Robust and Annotation-Efficient Plant Stress Phenotyping}
    
    \author[1]{Gurbhit Chaurakoti}
    \author[2]{Soumyashree Kar\corref{cor1}}

    \cortext[cor1]{Corresponding author. Email: soumyakar@iitb.ac.in}

    \affiliation[1]{organization={Department of Electrical Engineering, National Institute of Technology Delhi}, city={New Delhi}, postcode={110036}, country={India}}
    
    \affiliation[2]{organization={Centre of Studies in Resources Engineering, Indian Institute of Technology Bombay}, city={Powai, Mumbai}, postcode={400076}, country={India}}

\begin{abstract}
Semantic segmentation in agricultural imagery is often evaluated under in-domain protocols, yet practical deployment requires robustness to appearance perturbations, limited annotations, and cross-domain shift. This paper presents a diffusion-guided hybrid segmentation framework in which U-Net, DeepLabV3+, and SegFormer backbones generate coarse masks that are refined by Denoising Diffusion Probabilistic Models (DDPM), latent diffusion, or semantic-guided diffusion. The framework is evaluated through a 3$\times$3 architectural screening study on PlantSegV3, followed by boundary-constrained optimization, perturbation-guided retraining, low-data evaluation, constrained hyperparameter screening, and controlled cross-domain adaptation. On PlantSegV3, the best selected hybrid model achieves 71.83\% refined mean Intersection-over-Union (mIoU) and 26.10\% refined Boundary-F1, and the selected models remain stable under substantially reduced supervision, demonstrating strong annotation efficiency. Perturbation analysis identifies grayscale conversion, fog, coarse dropout, and shadow as the most disruptive appearance shifts, and the resulting augmentation policy substantially improves robustness during retraining. The adapted models further show effective transfer to external agricultural datasets under limited target supervision, indicating that diffusion refinement and boundary-aware optimization provide transferable structural priors. Overall, the results show that carefully matched backbone--refiner pairings, combined with perturbation-aware retraining, can improve structural delineation and robustness under realistic resource and distribution constraints.
\end{abstract}


\end{frontmatter}

\section{Introduction}
Semantic segmentation is a key step in plant phenotyping \citep{Karthiketal2022} as it enables pixel-level delineation and quantification of affected tissue, which is further used for downstream applications such as estimating lesion extent, spatial spread, disease burden and structural degradation \citep{Qinetal2025,kumar2026pixel}. In practical agricultural settings, however, images are affected by illumination changes, occlusion, blur, cluttered backgrounds, and large appearance variability across crops, acquisition devices, and field conditions \citep{app15169206}. These factors make stress segmentation substantially harder than benchmark performance alone suggests. A model that performs well in one dataset split may degrade sharply when the imaging conditions change or when only a small amount of labeled data is available. For this reason, robustness under perturbation and cross-domain shift is treated as a primary design objective in this study.\\
\\A second, equally important issue is computational efficiency. In many applied agricultural workflows, segmentation models must be developed and evaluated under limited hardware and training budgets rather than through exhaustive large-scale optimization. This is especially relevant when experiments are performed on local resources, where the available Graphics Processing Unit (GPU) memory and runtime constrain the number of architectures, training epochs, and hyperparameter configurations that can be explored \citep{rs14174217}. In this work, the entire model selection and retraining pipeline is designed around such constraints. The limited training budget is used as a realistic setting for testing whether diffusion-guided refinement can still provide measurable gains under resource-constrained conditions.
\\\\
Recent plant segmentation studies have improved in-domain performance by comparing Convolutional Neural Network (CNN), transformer, and hybrid backbones on benchmark datasets \citep{JIANG2024102144,POLLY2024100526,AGUSTIAN2026100357}. These works demonstrate that modern architectures can produce high-quality masks for diseased regions, but most are still evaluated under fixed training protocols and within the same domain. As a result, they provide limited insight into whether the learned masks remain useful under appearance shift, reduced supervision, or constrained training budgets. In plant phenotyping, this limitation is important because the final goal is not only accurate segmentation on a benchmark, but also reliable delineation of affected region in realistic field settings.
\\To avoid biasing the study toward a single inductive prior, we compare three representative segmentation backbones that span complementary design paradigms. U-Net, originally proposed by \citet{Ronnebergeretal2015} is a symmetric encoder–decoder with skip connections, originally designed to achieve precise localization from limited annotated data. \citet{Chenetal2018} introduced DeepLabV3+, which combines atrous spatial pyramid pooling with a lightweight decoder to merge multi-scale context and sharpen object boundaries. SegFormer represents a transformer-based alternative \citep{Xieetal2021}. It uses a hierarchically structured encoder with a lightweight Multi-Layer Perceptron (MLP) decoder, avoids positional encoding, and has shown strong robustness when the test resolution differs from the training resolution. Together, these three backbones cover local-detail, multi-scale-context, and global-attention design choices, making them a suitable basis for architectural screening under a fixed training budget.
\\In parallel with advances in segmentation architectures, diffusion models have recently emerged as powerful generative and refinement mechanisms for vision tasks \citep{Alimisisetal2025,10.1145/3626235}. In segmentation, they are increasingly explored as a way to improve structural coherence, suppress spurious predictions, and refine coarse masks \citep{WANG2025114481,electronics15071400,LIU2026100164} However, most existing formulations treat diffusion as a segmentation framework \citep{FANG2025103205} or as a mask-prior modelling mechanism \citep{Zhang2024PMSDiffPM,10.1609/aaai.v38i3.28068}, rather than as a carefully constrained refiner for plant imagery. This leaves open the question of which diffusion formulation best complements a given segmentation backbone, whether the resulting gains are stable across training regimes, and whether diffusion refinement remains useful when training must be done under limited computational resources.
\\ In this study, we compare three diffusion paradigms with different computational and conditioning assumptions. Denoising Diffusion Probabilistic Model (DDPM) is the canonical pixel-space denoising diffusion formulation and serves as the simplest stochastic refinement baseline \citep{Hoetal2020}. As presented by \citet{Rombachetal2022}, latent diffusion moves the denoising process into a learned latent space, substantially reducing computational cost while preserving controllability through conditioning. Recent semantically guided diffusion work shows that conditioning denoising on semantic priors can improve segmentation robustness and efficiency \citep{Liuetal2023}, motivating our semantic-guided refiner for plant imagery. Taken together, these refiners let us test whether gains come from diffusion itself, from latent-space efficiency, or from semantic conditioning.
\\It is also evident from existing literature that agricultural segmentation models frequently encounter performance degradation when tested under shifts in lighting, background complexity, sensor properties, or scene composition \citep{agronomy15010145,app15169206,Shafayetal2025}. Such shifts can reduce boundary fidelity and increase false positives or false negatives, especially for small or fragmented diseased regions. Robustness therefore cannot be inferred from a single in-domain score \citep{MAGISTRI2023108114}. A meaningful evaluation must also include perturbation sensitivity, recovery under augmentation-guided retraining, and behavior under reduced supervision\citep{WEI2025115051}. In this context, robustness is not only a generalization problem but also a practical requirement for deploying segmentation systems in the field \citep{biomedinformatics5020020}. Cross-domain transfer is especially challenging in plant imagery because datasets often differ in crop type, visual structure, annotation protocol, and foreground complexity \citep{MAGISTRI2023108114}. Few-shot adaptation can partially recover performance, but the extent of recovery depends on the source representation and the adaptation regime \citep{GE2025100121,PORTO2023100307}. A good segmentation model should therefore not only transfer reasonably well to a new dataset, but should also adapt efficiently when only a small amount of target supervision becomes available. This makes domain adaptation a natural extension of robustness analysis, particularly when the source model itself is trained and selected under constrained resources.
\\\\
Despite these advances, three gaps remain. First, most plant segmentation studies optimize for benchmark accuracy rather than robustness under perturbation, constrained training, and domain shift \citep{biomedinformatics5020020,WU20230038,MAGISTRI2023108114}. Second, diffusion-based segmentation is still underexplored as a refinement mechanism for stress-related plant masks, particularly when the target structure is small, irregular, or visually heterogeneous. Third, existing studies rarely connect architectural screening, boundary-aware optimization, perturbation analysis, controlled adaptation, and resource-aware model selection within a single unified framework. Consequently, there is still no clear answer to which backbone-refiner combinations are most stable, which perturbations matter most, how much performance can be recovered under limited target supervision, and how well such a pipeline can be executed under practical hardware constraints. 

The main contributions of this paper are as follows:
\begin{itemize}
    \item We propose a diffusion-guided hybrid segmentation framework for stress phenotyping in plant imagery, in which coarse masks generated by discriminative backbones are refined by DDPM, latent diffusion, or semantic-guided diffusion.
    \item We introduce a staged evaluation protocol that combines architectural screening, boundary-constrained optimization, perturbation sensitivity analysis, augmentation-guided retraining, low-data evaluation, and cross-domain transfer.
    \item We show that the proposed pipeline can be trained and selected under strict resource constraints on a local RTX 4050 6 Gigabytes (GB) Video Random Access Memory (VRAM) GPU, using short screening runs to identify promising backbone-refiner pairs before deeper evaluation.
    \item We demonstrate that diffusion refinement improves structural consistency when the backbone-refiner pairing is appropriate, while boundary-aware optimization further strengthens contour fidelity without destabilizing overlap-based performance.
    \item We show that the selected models remain stable under reduced supervision and that controlled target adaptation can recover a substantial portion of the performance lost under direct transfer.
    \item We further show that constrained training does not prevent useful model selection; instead, it provides a realistic setting for evaluating which architectural choices are genuinely robust and compute-efficient.
\end{itemize}

The remainder of the paper is organized as follows. Section 2 describes the datasets, framework, and training strategy. Section 3 presents the screening, optimization, robustness, and transfer results. Section 4 discusses the implications, limitations, and practical relevance of the proposed approach. Section 5 concludes the paper.

\section{Methodology}

\subsection{Experimental Setup}
The proposed study is designed as a staged evaluation pipeline for stress-related plant segmentation under limited computational resources. The central objective is to determine whether diffusion-guided refinement can improve the structural quality of coarse segmentation masks produced by discriminative backbones, and whether such gains persist under constrained training, perturbation, and cross-domain transfer. All source-domain experiments are conducted on PlantSegV3 dataset \citep{Weietal2026}, where the model search is intentionally restricted to a small number of backbone-refiner combinations and a short screening schedule in order to reflect realistic hardware limits. In particular, the full architecture grid is first screened using a compact training budget on a local RTX 4050 6 GB GPU, after which only the most promising configurations are retained for deeper optimization and robustness analysis. 

The evaluation protocol is organized into seven stages. First, a 3x3 architectural screening study is performed on the source dataset using a common optimization objective. Second, the selected configurations are retrained using a boundary-constrained objective to improve contour fidelity \citep{JIA2026}. Third, the same selected models are evaluated under reduced supervision to examine label efficiency \citep{Baranchuk2021LabelEfficientSS}. Fourth, a perturbation sensitivity analysis is carried out to identify the most disruptive appearance shifts and to derive an augmentation policy from the observed failure modes \citep{Prabhakaran2019PerturbationSA}. Fifth, the selected models are retrained using this augmentation policy under both full-data and reduced-data settings to examine robustness recovery. Sixth, constrained hyperparameter screening is carried out around the validated retraining regime \citep{WANG2021106602}. Seventh, the source-trained checkpoints are transferred to external datasets and adapted under controlled low-shot supervision to assess cross-domain generalization \citep{Rafietal2024}. 

Throughout the pipeline, both quantitative and qualitative outputs are recorded for the coarse prediction and the diffusion-refined output. Performance is assessed using mean Intersection-over-Union (mIoU), Dice score, pixel accuracy, and Boundary-F1. For each model, the refined prediction is treated as the primary output, while the coarse prediction is retained as the internal baseline. This allows the study to quantify not only absolute performance, but also the refinement gain introduced by the diffusion module.

\subsection{Datasets Used}
\begin{figure}[!t]
    \centering
    \includegraphics[width=\columnwidth]{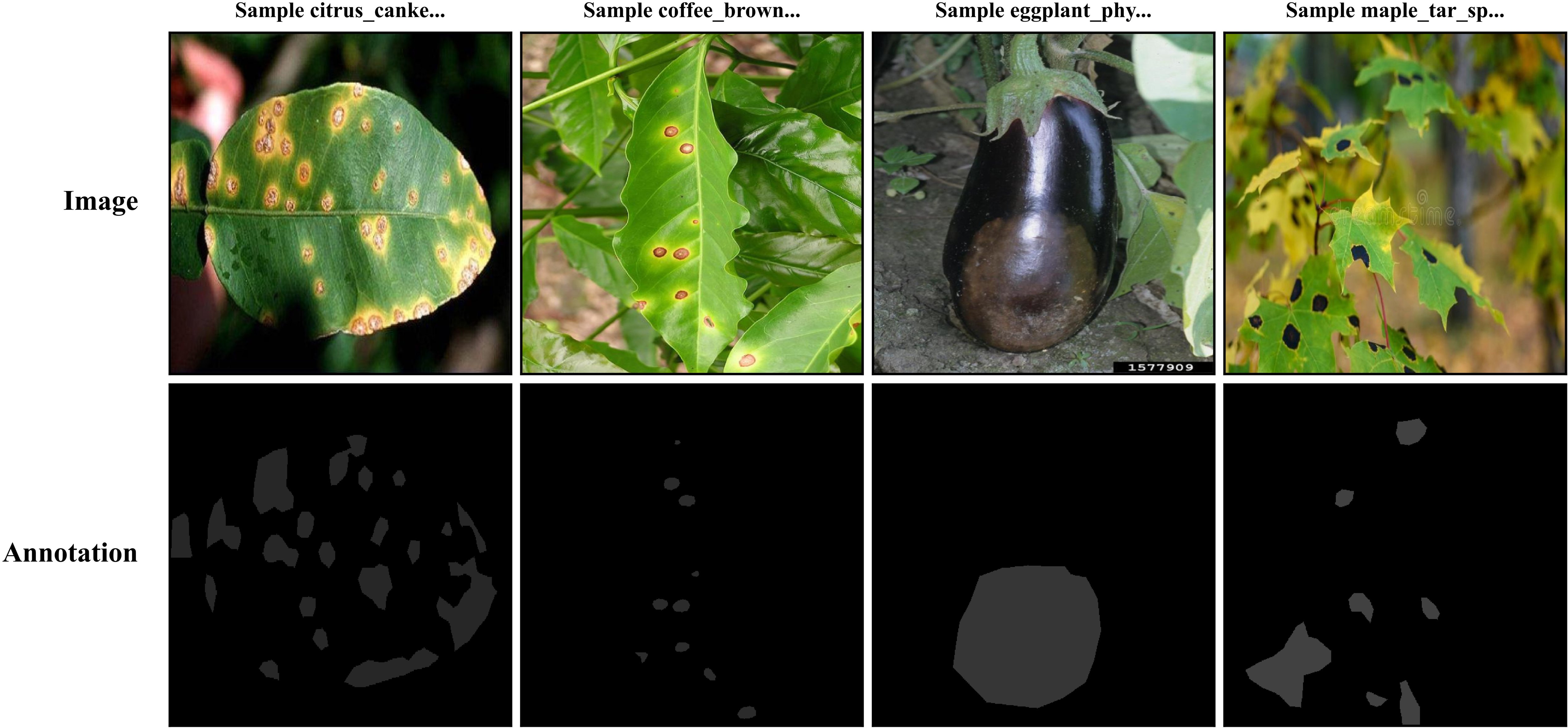}
    \caption{Qualitative segmentation samples from the PlantSegV3 dataset.}
    \label{fig:samples_plantsegv3}
\end{figure}

\begin{figure}[!t]
    \centering
    \includegraphics[width=\columnwidth]{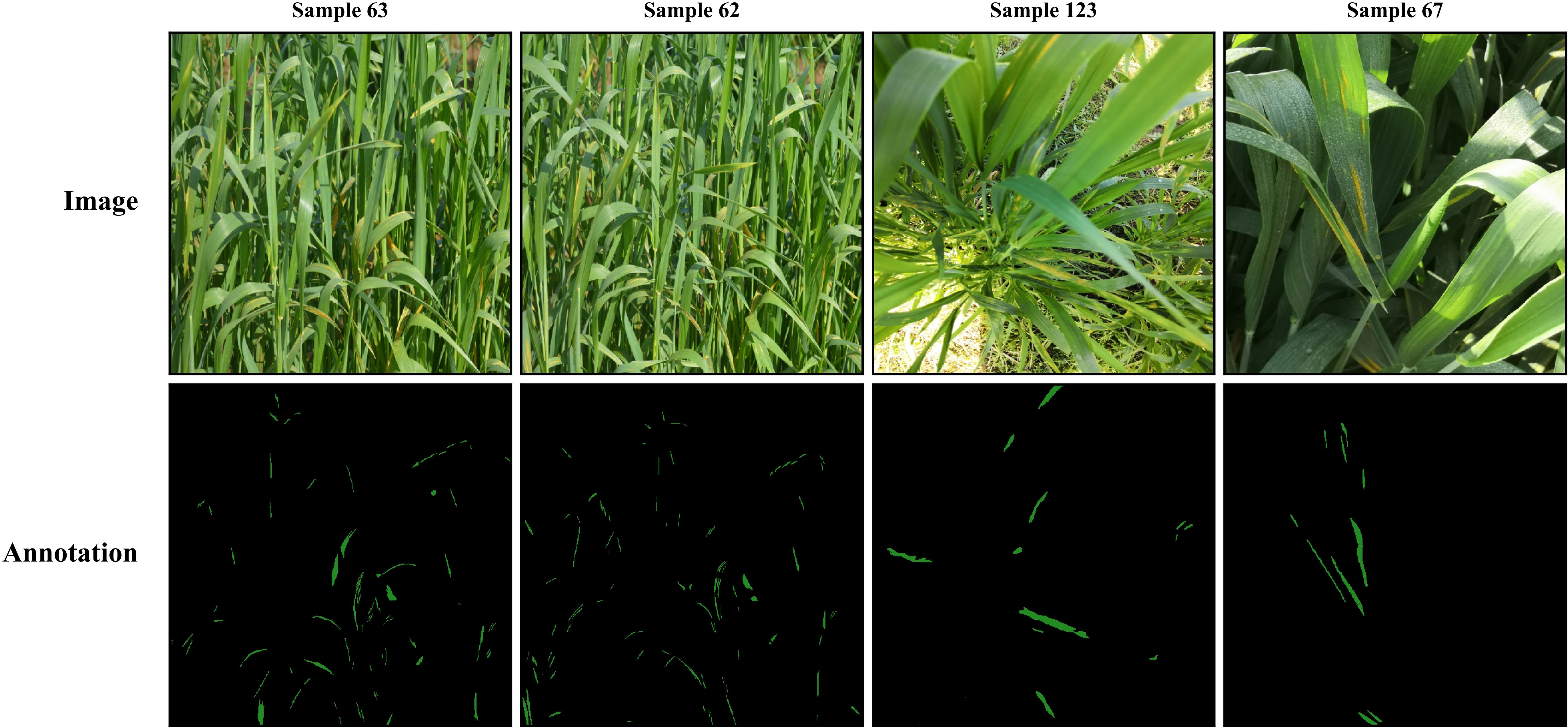}
    \caption{Qualitative segmentation samples from the NWRD dataset.}
    \label{fig:samples_nwrd}
\end{figure}

\begin{figure}[!t]
    \centering
    \includegraphics[width=\columnwidth]{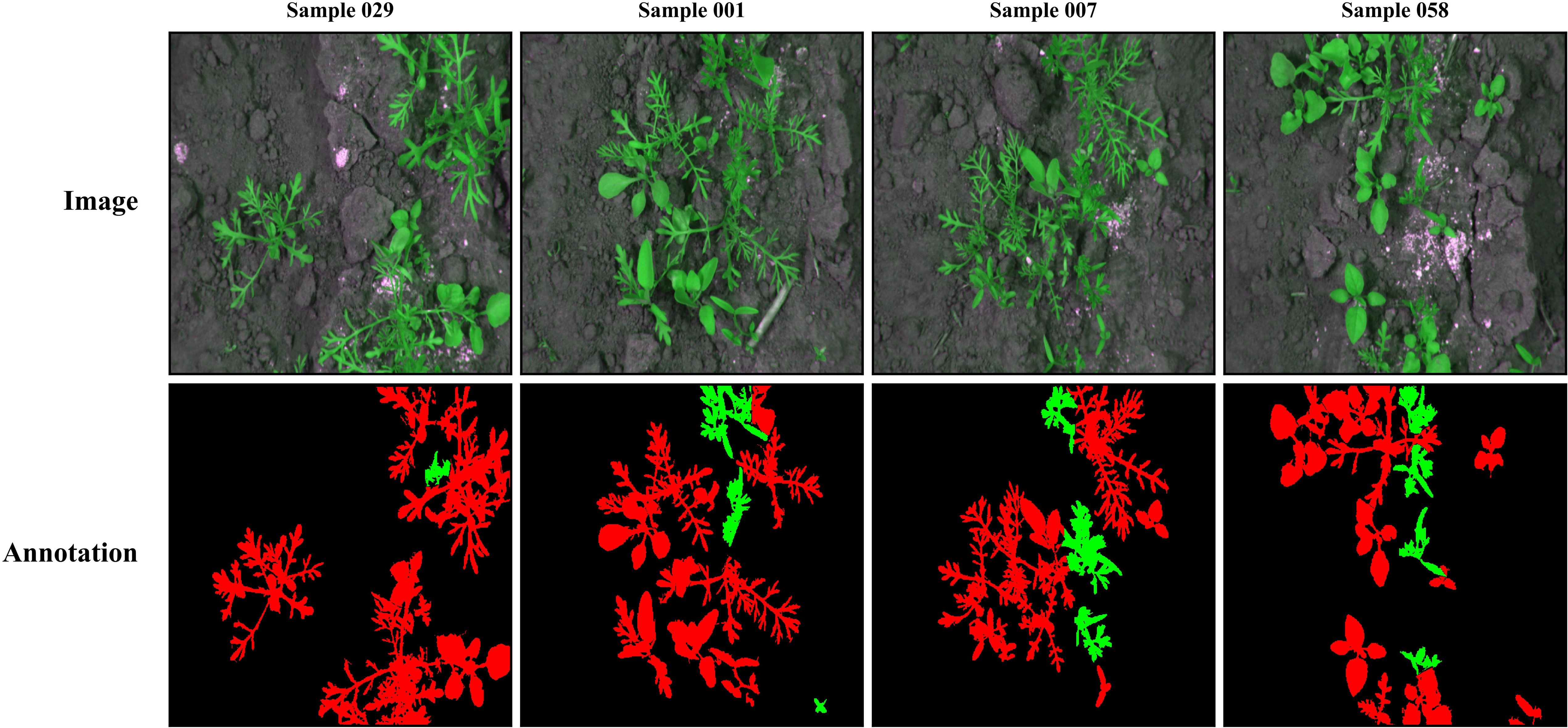}
    \caption{Qualitative segmentation samples from the CWFID dataset.}
    \label{fig:samples_cwfid}
\end{figure}

As summarized in Table~\ref{tab:dataset_summary}, three datasets are used in this study, which cover complementary transfer settings rather than a single homogeneous benchmark. PlantSegV3 (Figure~\ref{fig:samples_plantsegv3}) is used as the source-domain benchmark for architectural screening, boundary-aware optimization, low-data evaluation, perturbation-guided retraining, and constrained hyperparameter screening. The dataset provides in-the-wild plant disease segmentation masks and is large enough to support controlled comparison among hybrid segmentation configurations\citep{Weietal2026} In the original PlantSeg description, the dataset contains 7,774 diseased images across 34 host plants and 69 disease types, with 115 unique plant-disease cases and a standard 70/10/20 train/validation/test split. The annotation process uses pixel-level masks, and the dataset was curated specifically to support realistic disease segmentation under field conditions. The source benchmark in this work uses the PlantSegV3 train/validation split reported in the manuscript, containing 7,916 training image-mask pairs and 1,247 validation pairs.
\\\\NUST Wheat Rust Disease dataset (NWRD) (Figure~\ref{fig:samples_nwrd}) is used as an external transfer benchmark for disease-related localization and segmentation under a different acquisition context \citep{s23156942}. The dataset contains 100 high-resolution wheat-rust images, and the original benchmark reports 4000x6016 imagery that is divided into 224x224 patches to reduce memory and computation. Because the rust class is strongly imbalanced, the original study uses a 0.85/0.05/0.10 train/validation/test split together with augmentation and downsampling to balance the training set.\\
\\Crop/Weed Field Image Dataset (CWFID) (Figure~\ref{fig:samples_cwfid}) is used as a second transfer benchmark to assess whether the learned refinement behavior generalizes to another structured agricultural segmentation setting. The original CWFID dataset contains 60 top-down field images, with 162 crop plants and 332 weed plants, and the benchmark split uses 20 images for training and 40 for testing. Each image is annotated with a vegetation mask and crop/weed labels \citep{haug15}.

Figures~\ref{fig:samples_plantsegv3}--\ref{fig:samples_cwfid} illustrate the visual differences in foreground density, annotation structure, and background complexity. PlantSegV3 provides in-the-wild plant disease masks, NWRD captures high-resolution rust segmentation under a different acquisition context, and CWFID introduces structured crop-weed scenes with a different annotation ontology.

\begin{table*}[!t]
\centering
\caption{Summary of datasets used in this work.}
\label{tab:dataset_summary}
\small
\renewcommand{\arraystretch}{1.25}
\begin{tabularx}{\textwidth}{@{} l l c c >{\raggedright\arraybackslash}X >{\raggedright\arraybackslash}X @{}}
\toprule
\textbf{Dataset} & \textbf{Role in paper} & \textbf{Images} & \begin{tabular}{@{}c@{}}\textbf{ Benchmark Split} \\ \textbf{(Train/Val/Test)}\end{tabular} & \textbf{Annotation type} & \textbf{Main challenge} \\
\midrule
PlantSegV3 & Source benchmark & 7,774 & 0.70/0.10/0.20 & Pixel-wise disease masks & In-the-wild plant disease segmentation \\
NWRD & Transfer benchmark & 100 & 0.85/0.05/0.10 & Binary rust masks & High-resolution, imbalanced, patch-based images \\
CWFID & Transfer benchmark & 60 & 0.66/--/0.33 & Crop/weed semantic masks & Small-scale agricultural field scenes with semantic label-space shift \\
\bottomrule
\end{tabularx}
\end{table*}

\begin{figure*}[h]
\centering
\includegraphics[width=\textwidth]{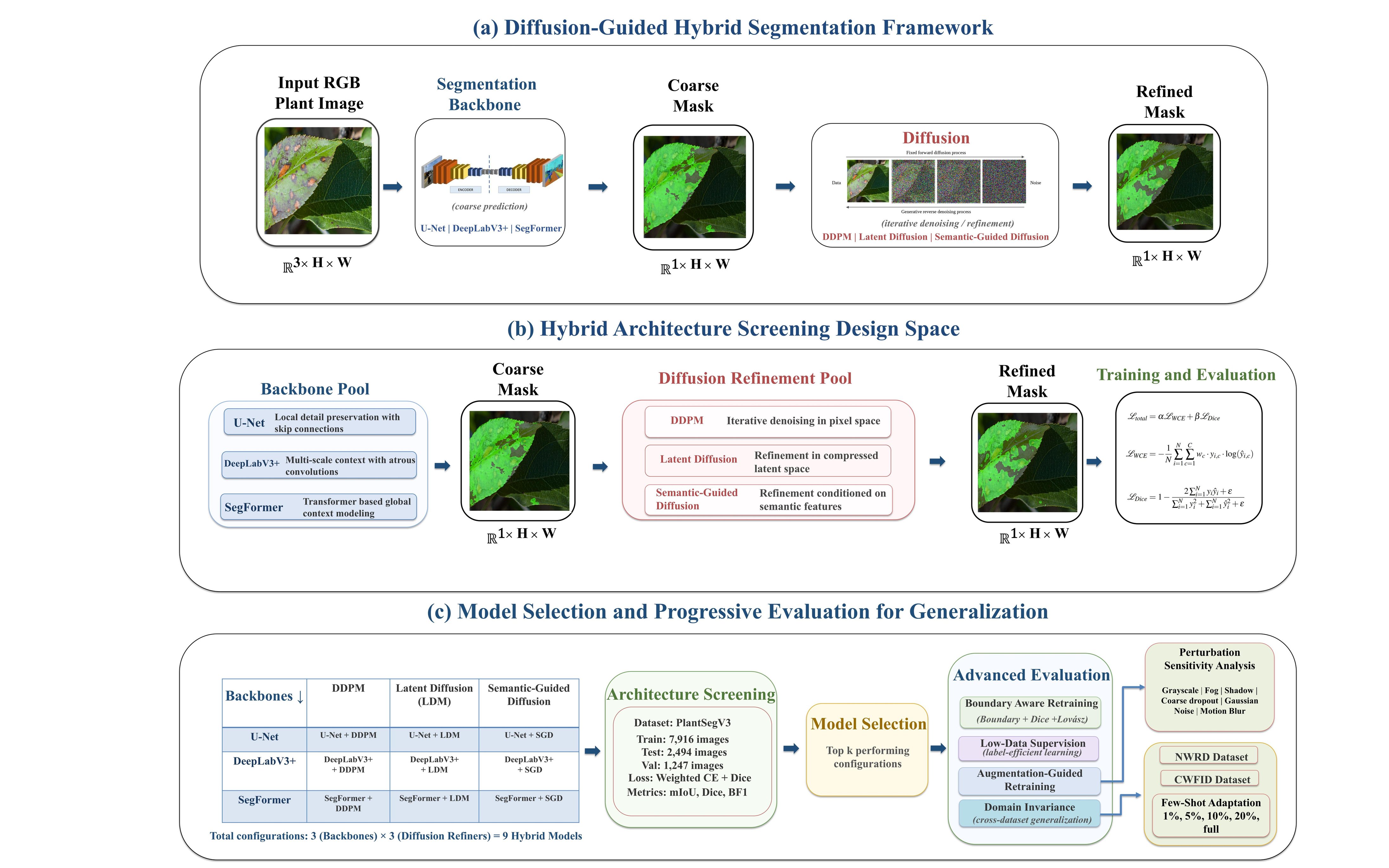}
\caption{Overview of the proposed diffusion-guided hybrid segmentation framework. 
The figure summarizes the coarse-to-refined prediction pipeline, the 3$\times$3 backbone--refiner screening space, and the subsequent evaluation stages for robustness, low-data learning, and cross-domain adaptation.
}\label{fig:framework}
\end{figure*}

\subsection{Diffusion-Guided Hybrid Segmentation Framework}
The proposed framework follows a coarse-to-refined design. Given an input image $X \in \mathbb{R}^{H \times W \times 3}$, a discriminative backbone first produces a coarse prediction map, which is then refined by a diffusion-based module. Let $f_b(\cdot)$ denote the backbone and $f_r(\cdot)$ denote the refiner. The coarse prediction is written as
\begin{equation}
P_c = f_b(X),
\end{equation}
and the refined prediction is written as
\begin{equation}
P_r = f_r(X, P_c, C),
\end{equation}
where $C$ denotes the conditioning signal used by the diffusion refiner. Depending on the configuration, $C$ may correspond to coarse logits, semantic guidance, or a compact representation derived from the backbone output.

Three backbone families are considered: U-Net, DeepLabV3+, and SegFormer. These architectures represent complementary segmentation paradigms spanning local detail preservation, multi-scale context aggregation, and transformer-based global representation learning \citep{Chenetal2018,Xieetal2021,Ronnebergeretal2015}. Three refinement strategies are paired with each backbone: DDPM, latent diffusion, and semantic-guided diffusion \citep{Hoetal2020,Rombachetal2022,Liuetal2023}. The combination of three backbones and three refiners produces a 3x3 design space of nine hybrid configurations. 

The role of diffusion in this framework is deliberately restricted to refinement. It is not used as a standalone segmentation predictor. Instead, the diffusion module acts as a structural correction mechanism that aims to improve boundary continuity, suppress spurious activations, and preserve coherent mask topology. This design makes the comparison across backbone-refiner pairs more interpretable, because the backbone remains responsible for the initial semantic localization while the diffusion branch focuses on refinement.
\begin{algorithm}[t]
\small
\caption{Staged evaluation of the proposed diffusion-guided hybrid segmentation framework}
\label{alg:staged_hybrid_pipeline}
\begin{algorithmic}[1]
\Require Source dataset $D_s$, backbone set $\mathcal{B}$, refiner set $\mathcal{R}$, low-shot fractions $\mathcal{F}$, perturbation set $\mathcal{T}$, target datasets $\mathcal{D}_t$
\Ensure Selected checkpoints and performance summaries

\State \textbf{Stage 1: Architectural screening.}
\State Train all $(b,r)\in\mathcal{B}\times\mathcal{R}$ on $D_s$ using $\mathcal{L}_{\mathrm{stage1}}$ and rank by validation mIoU and Boundary-F1.
\State Retain the top-$K$ configurations.

\State \textbf{Stage 2: Boundary-constrained optimization.}
\State Retrain the selected configurations using $\mathcal{L}_{\mathrm{bco}}$ and keep the best checkpoint per model.

\State \textbf{Stage 3: Low-data evaluation.}
\State Evaluate the selected models on nested fractions $\mathcal{F}$ of $D_s$.

\State \textbf{Stage 4: Perturbation sensitivity analysis.}
\State Measure performance drops under $\mathcal{T}$ and derive the augmentation policy $\mathcal{A}$.

\State \textbf{Stage 5: Augmentation-guided retraining.}
\State Retrain the selected models on $D_s$ using $\mathcal{A}$ and the boundary-aware objective.

\State \textbf{Stage 6: Constrained hyperparameter screening.}
\State Tune a compact neighborhood around the validated retraining regime.

\State \textbf{Stage 7: Cross-domain transfer.}
\State Adapt the selected checkpoints to each $D_t\in\mathcal{D}_t$ under low-shot supervision.

\State \Return Selected checkpoints and evaluation results
\end{algorithmic}
\end{algorithm}

\subsection{Training Strategy}
The training strategy is organized into progressively stricter stages, as summarized in Algorithm~\ref{alg:staged_hybrid_pipeline}. In Stage 1, all nine backbone-refiner combinations are screened under a fixed optimization objective. The same loss function is used for every configuration to ensure that differences in performance are driven by architecture rather than by training recipe. The Stage-1 objective combines cross-entropy and Dice loss \citep{YEUNG2022102026}:
\begin{equation}
\mathcal{L}_{\mathrm{stage1}} = \lambda_{\mathrm{ce}} \mathcal{L}_{\mathrm{ce}} + \lambda_{\mathrm{dice}} \mathcal{L}_{\mathrm{dice}}.
\end{equation}
This stage is intentionally short and serves as a computationally efficient filter for identifying stable hybrid models from the diverse candidate configurations presented in Table~\ref{tab:model_scenarios}. These configurations encompass distinct paradigms of diffusion refinement integrated across three standard backbone architectures. The output of this stage is not the final conclusion, but a ranking of candidate configurations based on validation performance, refinement gain, and convergence stability. 

In Stage 2, the top-performing configurations are retrained using a boundary-sensitive objective. This stage is intended to improve contour fidelity without sacrificing the region-level quality achieved during screening. The selected models are optimized further only after Stage-1 screening has identified which backbone-refiner pairs are worth deeper refinement. 

In Stage 3, the selected models are evaluated under reduced supervision to examine label efficiency. The training subsets are progressively reduced so that the effect of annotation scarcity can be isolated from the effect of architectural change. This stage is important for assessing whether the selected hybrid architectures preserve performance when supervision is limited. 

In Stage 4, the selected models are subjected to perturbation-guided analysis. The perturbation ranking obtained in this stage is then used to define the augmentation policy for retraining. 

In Stage 5, the selected models are retrained using the perturbation-derived augmentation policy. The retraining proceeds on PlantSegV3 only, while the validation split remains clean, so that robustness is learned from source-domain failure modes rather than from target labels. 

In Stage 6, a constrained hyperparameter screening is performed around the validated retraining regime. Rather than exploring a large search space, the tuning is restricted to a narrow set of optimization parameters so that the overall pipeline remains resource-aware. 

In Stage 7, the source-domain checkpoints are transferred to the external benchmarks and adapted under controlled supervision. This final stage tests whether the source-domain refinement behavior can be recovered or preserved under domain shift.
\begin{table*}[!t]
\centering
\small
\caption{Summary of model scenarios used in the comparative evaluation. A tick indicates the presence of a component and a cross indicates its absence. DDPM, LDM (Latent Diffusion Model), and Semantic-Guided Diffusion (SGD) denote mask-space diffusion, latent diffusion, and semantic-guided diffusion, respectively. All nine configurations are screened with a weighted composite objective combining weighted cross-entropy and weighted Dice loss, as shown in the last column.}
\label{tab:model_scenarios}
\resizebox{\textwidth}{!}{%
\begin{tabular}{c l c l c c c c c c l}
\toprule
ID & Model category & Code & Name & Local detail & Multi-scale context & Global context & Iterative denoising & Latent compression & Semantic conditioning & Loss \\
\midrule

\multirow{3}{*}{1} & \multirow{3}{*}{U-Net hybrids} 
& 1-1 & U-Net + DDPM & \cmark & \xmark & \xmark & \cmark & \xmark & \xmark & $L_{\mathrm{WCE+WDice}}$ \\
& & 1-2 & U-Net + LDM  & \cmark & \xmark & \xmark & \cmark & \cmark & \xmark & $L_{\mathrm{WCE+WDice}}$ \\
& & 1-3 & U-Net + SGD  & \cmark & \xmark & \xmark & \cmark & \xmark & \cmark & $L_{\mathrm{WCE+WDice}}$ \\
\midrule

\multirow{3}{*}{2} & \multirow{3}{*}{DeepLabV3+ hybrids} 
& 2-1 & DeepLabV3+ + DDPM & \cmark & \cmark & \xmark & \cmark & \xmark & \xmark & $L_{\mathrm{WCE+WDice}}$ \\
& & 2-2 & DeepLabV3+ + LDM  & \cmark & \cmark & \xmark & \cmark & \cmark & \xmark & $L_{\mathrm{WCE+WDice}}$ \\
& & 2-3 & DeepLabV3+ + SGD  & \cmark & \cmark & \xmark & \cmark & \xmark & \cmark & $L_{\mathrm{WCE+WDice}}$ \\
\midrule

\multirow{3}{*}{3} & \multirow{3}{*}{SegFormer hybrids} 
& 3-1 & SegFormer + DDPM & \xmark & \cmark & \cmark & \cmark & \xmark & \xmark & $L_{\mathrm{WCE+WDice}}$ \\
& & 3-2 & SegFormer + LDM  & \xmark & \cmark & \cmark & \cmark & \cmark & \xmark & $L_{\mathrm{WCE+WDice}}$ \\
& & 3-3 & SegFormer + SGD  & \xmark & \cmark & \cmark & \cmark & \xmark & \cmark & $L_{\mathrm{WCE+WDice}}$ \\
\bottomrule
\end{tabular}%
}

\end{table*}

\begin{figure*}[h]
    \centering
    \includegraphics[width=\textwidth]{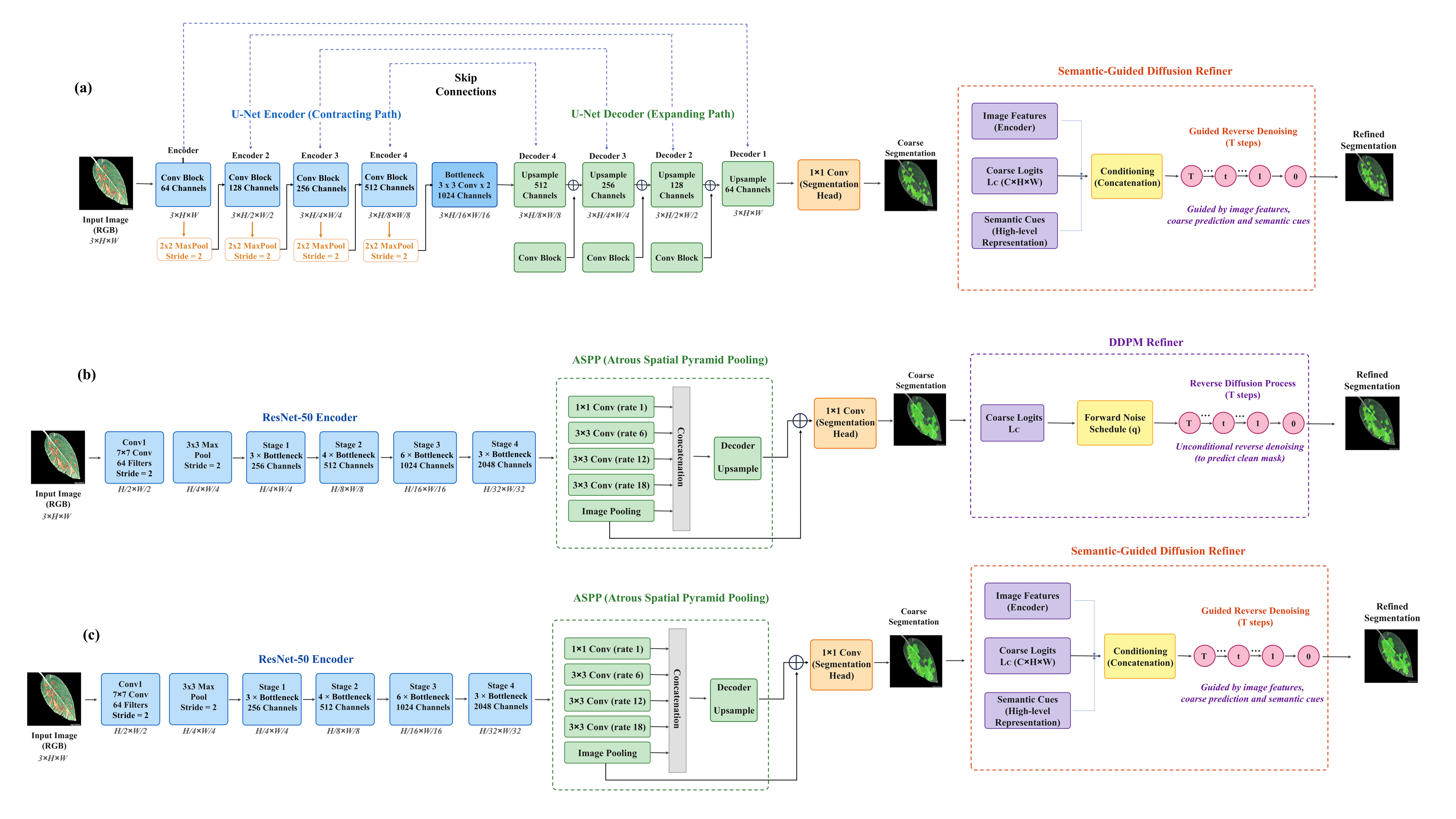}
    \caption{Architectures of the three selected hybrid segmentation configurations used in this study: 
(a) U-Net + semantic-guided diffusion, 
(b) DeepLabV3+ + DDPM, and 
(c) DeepLabV3+ + semantic-guided diffusion. 
Each subfigure shows the backbone encoder, the prediction head, and the corresponding diffusion-based refinement branch.}
\label{fig:model_architecture}
\end{figure*}

\subsection{Boundary-Constrained Optimization}
Boundary-constrained optimization is applied to the selected Stage-1 models to strengthen structural delineation. This stage is motivated by the observation that overlap-based metrics alone do not sufficiently capture the quality of thin, irregular, or fragmented regions. To address this, the objective combines three complementary terms:
\begin{equation}
\mathcal{L}_{\mathrm{bco}} = \lambda_1 \mathcal{L}_{\mathrm{dice}} + \lambda_2 \mathcal{L}_{\mathrm{boundary}} + \lambda_3 \mathcal{L}_{\mathrm{lovasz}}.
\end{equation}
The Dice term encourages overlap, the boundary term penalizes contour mismatch, and the Lovasz component improves region-level optimization under imbalanced foreground structure. In implementation, the boundary term is computed from gradient-aligned differences between the predicted mask and the ground truth mask, so that the network is explicitly penalized when the predicted boundary deviates from the reference boundary \citep{Terven2025Comprehensive}. The selected checkpoints are chosen by minimum validation loss under the composite objective, with refined mIoU and Boundary-F1 used as secondary indicators of stability. This stage is important because it allows the diffusion refiner to be evaluated not only as a coarse denoiser, but as a structural correction mechanism that is sensitive to contour quality.

\subsection{Low-Data Evaluation under Reduced Supervision}
After boundary-constrained optimization, the selected models are evaluated under progressively reduced supervision. Smaller labeled subsets are sampled in a nested manner from the training set, while the validation split is kept fixed. This design isolates the effect of label scarcity and allows the study to determine whether the selected hybrid architectures preserve performance when supervision is limited \citep{Fawakherji2026Learning}.

The low-data experiment is important for two reasons. First, it tests whether the diffusion refiner and the boundary-aware objective remain stable when the model sees fewer annotated examples. Second, it provides evidence for whether the proposed framework can support practical deployment scenarios in which extensive annotation is not feasible. Performance is reported using the same metrics as in the full-data setting so that the effect of reduced supervision can be compared directly with the source-domain baseline.

\subsection{Perturbation Sensitivity Analysis}
To identify which appearance changes most strongly affect segmentation quality, a perturbation sensitivity analysis is performed on the source-domain model. The analysis includes grayscale conversion, fog, coarse dropout, shadow, Gaussian noise, blur, motion blur, brightness shifts, contrast shifts, gamma shifts, affine transformations, perspective distortion, and elastic deformation. Each perturbation is applied independently in a deterministic evaluation setting, and the resulting refined prediction is compared with the clean baseline \citep{tavera2022augmentation}. For each perturbation, the drop in mIoU and Boundary-F1 relative to the clean condition is measured. The perturbations are then ranked by severity, allowing the training pipeline to identify which distortions are most damaging to structural segmentation. This ranking is used to derive the augmentation policy for the next stage. In this way, the retraining process becomes failure-mode driven rather than manually tuned.

\subsection{Augmentation-Guided Retraining}
The perturbation ranking is converted into an augmentation policy that is used for retraining the selected models on PlantSegV3. The policy is deliberately structured in two phases. The first phase applies mild geometric and photometric variation to maintain stability while avoiding excessive regularization. The second phase introduces stronger perturbations that reflect the most severe failure modes identified in the sensitivity analysis. This includes appearance shifts such as shadowing, haze, blur, noise, contrast variation, and partial occlusion. 

Retraining is performed only on the source domain, while the validation split remains clean. This design ensures that the augmentation policy is learned as a robustness mechanism rather than as a substitute for target supervision. The same boundary-constrained loss used in Stage 2 is retained during this retraining stage so that structural fidelity and overlap quality are optimized jointly. Because the models are trained under a small local hardware budget, the retraining schedule is intentionally compact. The goal is not to maximize training duration, but to test whether a carefully selected augmentation policy can improve robustness without requiring large-scale compute.

\subsection{Constrained Hyperparameter Screening}
Hyperparameter tuning is conducted in a constrained manner. Rather than performing exhaustive search over large grids, a compact set of hyperparameters is used to identify stable operating points for the selected models. 

The constrained screening philosophy is consistent with the overall goal of the paper: to evaluate whether diffusion-guided segmentation can remain effective under practical training limits \citep{RAIAAN2024100470}. Accordingly, hyperparameters are treated as fixed experimental settings after selection, not as a separate optimization target.

\subsection{Cross-Domain Generalization and Controlled Adaptation}
Finally, the selected source-domain checkpoints are transferred to NWRD and CWFID to study cross-domain generalization \citep{Rafietal2024}. Direct transfer is first evaluated without target-side fine-tuning. This zero-shot evaluation measures how much of the learned structural bias remains intact when the model is exposed to a new domain. The same checkpoints are then adapted using small labeled target subsets under a controlled low-shot protocol. The adaptation protocol is designed to answer a practical question: how much of the lost performance can be recovered once a small amount of target supervision becomes available? In this setting, the emphasis is not on exhaustive domain adaptation engineering, but on controlled recovery of structural quality. This makes the adaptation study directly relevant to field deployment, where annotated target data are often scarce and expensive to obtain.

Overall, the methodology is deliberately resource-aware, robustness-oriented, and evaluation-driven. The study is designed to answer not only which backbone-refiner combination performs best, but also which combinations are stable, which perturbations matter most, and how well the resulting pipeline can operate under realistic training constraints.

\subsection{Evaluation Metrics and Optimization Objective}
The proposed framework is evaluated using a complementary set of region-based, boundary-based, and pixel-wise metrics. This is necessary because stress segmentation quality is not determined only by the amount of overlap between the predicted and reference masks, but also by how well the model preserves thin structures, irregular boundaries, and small disconnected regions. Accordingly, mean Intersection-over-Union (mIoU) and Dice score quantify region-level agreement, Boundary-F1 measures contour fidelity, and pixel accuracy reports the fraction of correctly labeled pixels over the full image \citep{MOHAMMADI2025100624,Mohamed2021IncorporationOR,bf1}. Beyond delineation, these metrics provide pixel-level quantification of the affected region, allowing the model to be evaluated in terms of how much of the target structure is correctly recovered rather than only whether the disease is detected at the image level.

Let $Y$ and $\hat{Y}$ denote the ground-truth and predicted masks, respectively, and let $C$ be the number of classes. For class $c$, let $TP_c$, $FP_c$, and $FN_c$ denote the true positives, false positives, and false negatives. The class-wise Intersection-over-Union is defined as
\begin{equation}
\mathrm{IoU}_c = \frac{TP_c}{TP_c + FP_c + FN_c}.
\end{equation}
The mean Intersection-over-Union is then computed as
\begin{equation}
\mathrm{mIoU} = \frac{1}{C}\sum_{c=0}^{C-1} \mathrm{IoU}_c.
\end{equation}

The Dice coefficient is defined as
\begin{equation}
\mathrm{Dice}_c = \frac{2TP_c}{2TP_c + FP_c + FN_c},
\end{equation}
with the macro-averaged Dice score given by
\begin{equation}
\mathrm{Dice} = \frac{1}{C}\sum_{c=0}^{C-1} \mathrm{Dice}_c.
\end{equation}

Pixel accuracy is computed as
\begin{equation}
\mathrm{PA} = \frac{1}{N}\sum_{i=1}^{N}\mathbb{I}\!\left(\hat{y}_i = y_i\right),
\end{equation}
where $N$ is the total number of pixels and $\mathbb{I}(\cdot)$ is the indicator function.

To assess boundary quality, we extract the boundary map $\mathcal{B}(\cdot)$ of a mask using morphological erosion and define a dilated boundary operator $\delta(\cdot)$ for tolerance matching. Boundary precision and boundary recall are then computed from the overlap between the predicted and ground-truth boundaries, and their harmonic mean gives Boundary-F1:
\begin{equation}
\mathrm{BF1} = \frac{2 \cdot P_b \cdot R_b}{P_b + R_b},
\end{equation}
where $P_b$ and $R_b$ are the boundary precision and boundary recall, respectively. In this work, BF1 is reported because it is more sensitive than overlap-based scores to contour sharpness and boundary leakage, which are especially important for fragmented stress regions.
\\\\Boundary-F1 is computed with a one-pixel tolerance band around the ground-truth contour. A boundary prediction is therefore counted as correct only when it lies within this narrow neighborhood, which makes the metric intentionally strict. As a result, small contour shifts, tiny holes, or minor boundary leakage can reduce the score even when the overlap-based metrics remain high. This is desirable when the application requires sharp delineation, but it also means BF1 is often lower than mIoU or Dice for thin and fragmented stress regions. Boundary-sensitive evaluation studies have shown that such metrics reveal contour errors that standard mask-overlap scores can overlook, while tolerant segmentation metrics explicitly use the tolerance radius to define acceptable boundary deviations \citep{Mohamed2021IncorporationOR,bf1}.
\\\\During training, the selected models are optimized with a composite objective that combines weighted cross-entropy, Dice loss, boundary loss, and Lov\'asz loss. Weighted cross-entropy addresses class imbalance by penalizing misclassification of minority foreground pixels more strongly, Dice loss improves overlap under skewed foreground-background ratios, boundary loss explicitly encourages sharper contours, and Lov\'asz loss serves as a differentiable surrogate aligned with IoU optimization. The overall objective is written as
\begin{equation}
\begin{aligned}
\mathcal{L}_{\mathrm{total}}
=&\;
\lambda_{\mathrm{wce}}\mathcal{L}_{\mathrm{wce}}
+\lambda_{\mathrm{dice}}\mathcal{L}_{\mathrm{dice}}\\
&+\lambda_{\mathrm{boundary}}\mathcal{L}_{\mathrm{boundary}}
+\lambda_{\mathrm{lovasz}}\mathcal{L}_{\mathrm{lovasz}}.
\end{aligned}
\end{equation}

This composite formulation is used throughout the staged evaluation pipeline, with the loss weights kept fixed after selection so that the observed differences in performance arise primarily from the architecture, refinement strategy, and training regime rather than from the optimization recipe.

\section{Results and Discussion}

This section evaluates the proposed diffusion-guided hybrid framework in the same order as the methodology: architectural screening, boundary-constrained optimization, low-data evaluation, perturbation sensitivity analysis, augmentation-guided retraining, constrained hyperparameter screening, and cross-domain transfer. The results show that diffusion refinement is most useful when it is treated as a structural correction stage rather than as a replacement for the segmentation backbone. They also show that the proposed pipeline remains effective under a compact training budget, which is important because all source-domain experiments are conducted under practical compute constraints.

\subsection{Architectural Screening on PlantSegV3}
The initial 3x3 screening study identifies clear architecture-dependent behavior under the shared Stage-1 objective. Among the nine backbone--refiner combinations, DeepLabV3+ + DDPM achieves the strongest validation performance with a refined mIoU of 0.7043, followed closely by U-Net + semantic-guided diffusion at 0.7026 and U-Net + DDPM at 0.6998. DeepLabV3+ + semantic-guided diffusion yielded the most substantial positive refinement deltas among all competitive models, producing an mIoU improvement of +0.0128 and a Boundary-F1 increase of +0.0120 over its coarse backbone predictions. In contrast, both latent-diffusion variants fail to provide a consistent improvement over the corresponding coarse predictions, and the SegFormer branch remains substantially weaker, with refined mIoU values in the 0.5049--0.5195 range.

The coarse-versus-refined comparison is important because the framework is designed around refinement rather than replacement. In the best-performing configurations, diffusion produces a measurable but modest improvement over the backbone output. DeepLabV3+ + DDPM improves from 0.6918 coarse mIoU to 0.7043 refined mIoU and increases Boundary-F1 from 0.0644 to 0.0755. U-Net + semantic-guided diffusion shows a similar gain, improving from 0.6900 to 0.7026 in mIoU and from 0.0750 to 0.0797 in Boundary-F1. These gains confirm that the diffusion branch adds useful structural correction when the backbone--refiner pairing is appropriate.

\begin{figure}[!t]
    \centering
    \includegraphics[width=\columnwidth]{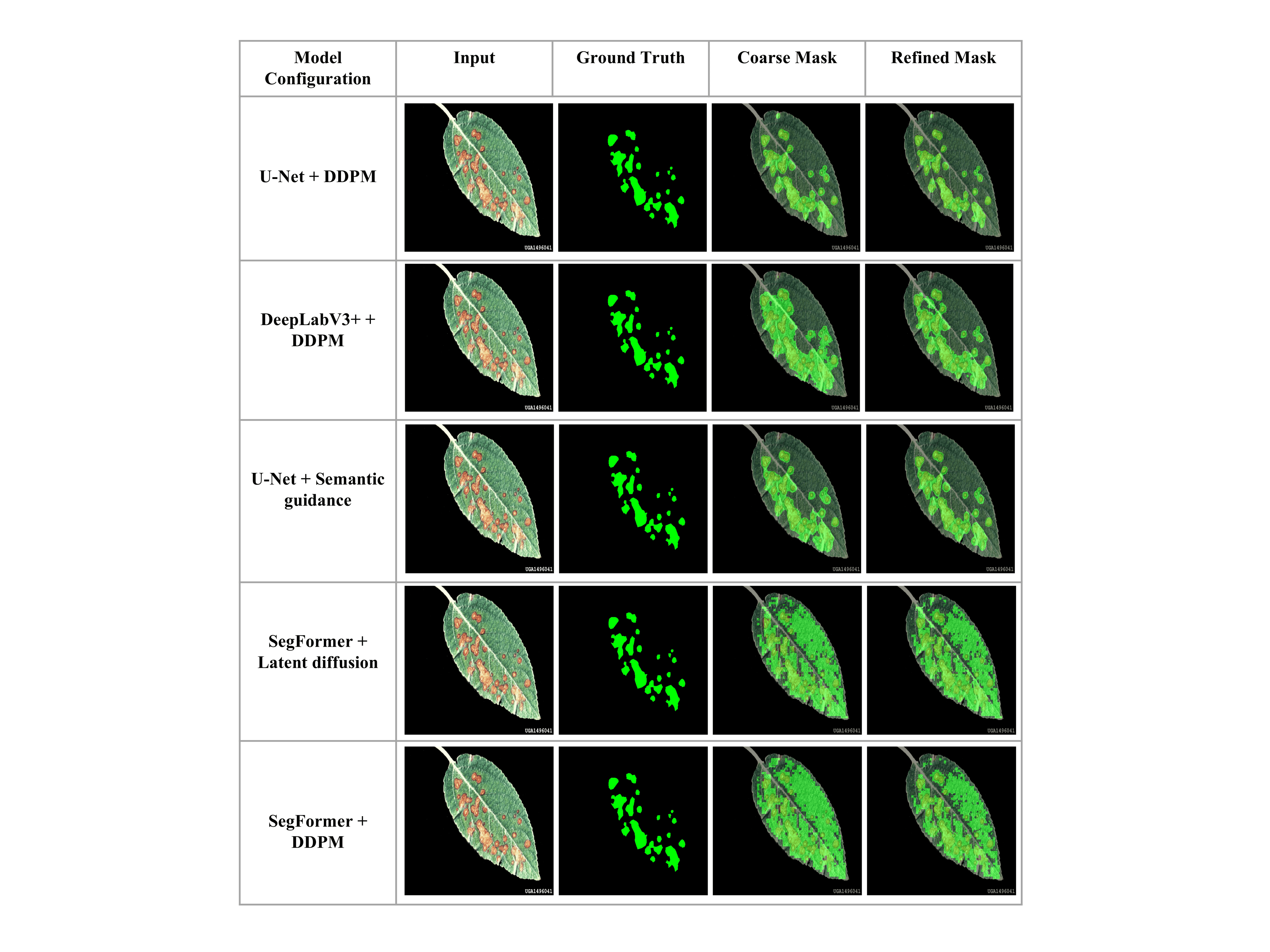}
    \caption{
Qualitative comparison of coarse and diffusion-refined predictions, where the same input sample is tested across different models to facilitate direct comparison of refinement behavior. Quantitative screening results are reported in Table ~\ref{tab:full_metrics_summary}.
}
\label{fig:qualitative_panels}
\end{figure}

\begin{figure}[!t]
    \centering
    \includegraphics[width=\columnwidth]{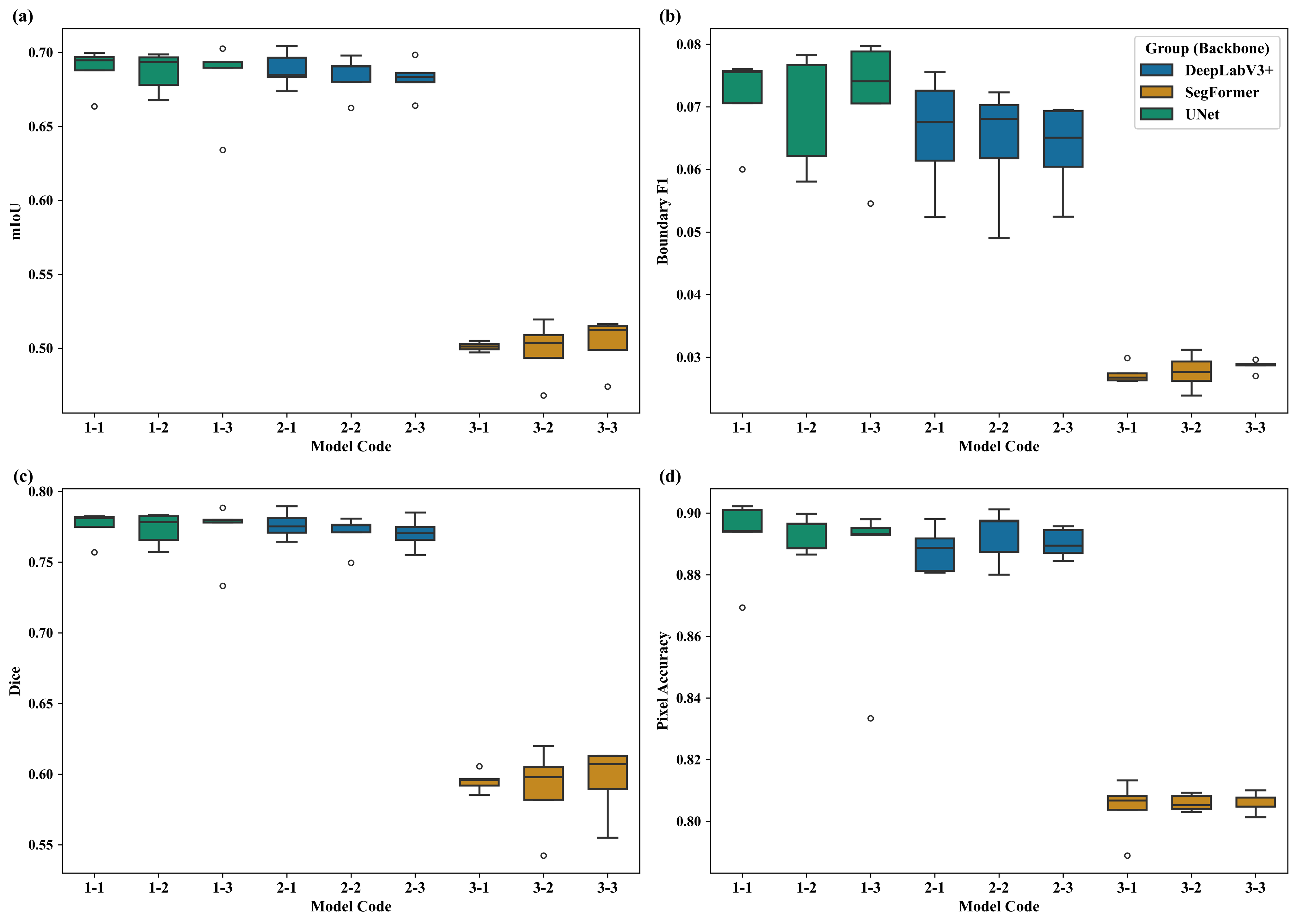}
    \caption{Performance comparison of all screened hybrid configurations on the PlantSegV3 dataset. Box plots illustrate the distribution of validation metrics for the nine backbone-refiner combinations. Panel (a) displays refined mIoU, panel (b) shows refined Boundary-F1, panel (c) presents the Dice coefficient, and panel (d) illustrates Pixel Accuracy. Colors denote the base architecture group, and the x-axis codes correspond to the specific configurations detailed in Table~\ref{tab:full_metrics_summary}.}
    \label{fig:metrics_stage1}
\end{figure}

Overall, the screening stage suggests three practical conclusions. First, the convolutional backbones are more stable than the transformer branch under the current short-screening regime. Second, DDPM and semantic-guided diffusion are the most reliable refiners in this setting. Third, latent diffusion is not consistently beneficial when training is intentionally brief and compute-limited. This is not a failure of the framework; rather, it indicates that diffusion priors must be matched carefully to the backbone representation and the available training budget. Table~\ref{tab:full_metrics_summary} reports the coarse and refined validation metrics for all nine backbone-refiner combinations, whereas Figure~\ref{fig:qualitative_panels} shows representative coarse-to-refined predictions and Figure~\ref{fig:metrics_stage1} summarizes the distribution of validation metrics across the screened configurations. Taken together, these results show that the effect of diffusion refinement is architecture-dependent rather than uniform across all pairings.

Out of the nine evaluated backbone-refiner combinations, we selected DeepLabV3+ with DDPM, U-Net with semantic-guided diffusion, and DeepLabV3+ with semantic-guided diffusion as our primary architectures. The full 3$\times$3 design space is summarized in Table~\ref{tab:model_scenarios}, while the layer-level layouts of the three selected backbone-refiner configurations are shown in Figure~\ref{fig:model_architecture}. The figure is included to clarify how the backbone, prediction head, and diffusion-based refiner interact in each retained model.

\begin{sidewaystable}[!htbp]
\centering
\caption{Stage-1 3x3 screening on PlantSegV3. For each configuration, the best validation epoch within the 5-epoch screening run is reported. Qualitative comparison of coarse and refined masks reported in Figure ~\ref{fig:qualitative_panels}.}
\label{tab:full_metrics_summary}
\setlength{\tabcolsep}{4.5pt} 
\begin{tabular}{llccccccccccccc} 
\toprule
\multirow{2}{*}{\textbf{Code}} & \multirow{2}{*}{\textbf{Architecture}} & \multirow{2}{*}{\textbf{Refiner}} & \multicolumn{3}{c}{\textbf{mIoU}} & \multicolumn{3}{c}{\textbf{Boundary F1}} & \multicolumn{3}{c}{\textbf{Dice}} & \multicolumn{3}{c}{\textbf{Pixel Accuracy}} \\
\cmidrule(lr){4-6} \cmidrule(lr){7-9} \cmidrule(lr){10-12} \cmidrule(lr){13-15}
& & & Coarse & Refined & $\Delta$ & Coarse & Refined & $\Delta$ & Coarse & Refined & $\Delta$ & Coarse & Refined & $\Delta$ \\
\midrule
1-1 & UNet & DDPM & 0.6969 & 0.6998 & +0.0028 & 0.0761 & 0.0761 & +0.0000 & 0.7836 & 0.7825 & -0.0011 & 0.8913 & 0.9023 & +0.0110 \\
1-2 & UNet & Latent Diff & 0.7016 & 0.6988 & -0.0029 & 0.0805 & 0.0783 & -0.0022 & 0.7874 & 0.7833 & -0.0040 & 0.8961 & 0.8998 & +0.0037 \\
1-3 & UNet & Semantic & 0.6900 & 0.7026 & +0.0126 & 0.0750 & 0.0797 & +0.0047 & 0.7800 & 0.7885 & +0.0086 & 0.8830 & 0.8981 & +0.0151 \\
\addlinespace
2-1 & DeepLabV3+ & DDPM & 0.6918 & 0.7043 & +0.0126 & 0.0644 & 0.0755 & +0.0111 & 0.7820 & 0.7896 & +0.0076 & 0.8806 & 0.8981 & +0.0175 \\
2-2 & DeepLabV3+ & Latent Diff & 0.6986 & 0.6980 & -0.0007 & 0.0706 & 0.0723 & +0.0018 & 0.7849 & 0.7808 & -0.0041 & 0.8937 & 0.9013 & +0.0075 \\
2-3 & DeepLabV3+ & Semantic & 0.6856 & 0.6984 & +0.0128 & 0.0574 & 0.0695 & +0.0120 & 0.7773 & 0.7852 & +0.0078 & 0.8816 & 0.8958 & +0.0141 \\
\addlinespace
3-1 & SegFormer & DDPM & 0.5019 & 0.5049 & +0.0030 & 0.0270 & 0.0299 & +0.0029 & 0.6064 & 0.6056 & -0.0007 & 0.7783 & 0.7888 & +0.0106 \\
3-2 & SegFormer & Latent Diff & 0.4922 & 0.5195 & +0.0273 & 0.0239 & 0.0312 & +0.0073 & 0.5837 & 0.6200 & +0.0363 & 0.8064 & 0.8030 & -0.0034 \\
3-3 & SegFormer & Semantic & 0.4999 & 0.5164 & +0.0165 & 0.0244 & 0.0289 & +0.0044 & 0.5892 & 0.6130 & +0.0237 & 0.8127 & 0.8077 & -0.0049 \\
\bottomrule
\end{tabular}
\end{sidewaystable}

\subsection{Boundary-Constrained Optimization for Structural Delineation}
After screening, the selected configurations are retrained using the boundary-constrained objective composed of Dice, boundary alignment, and Lovasz terms. This stage improves structural delineation without destabilizing the region-level segmentation quality obtained during screening. The optimized checkpoints remain tightly grouped in performance, with refined mIoU values of 0.7114 for DeepLabV3+ + DDPM, 0.7052 for DeepLabV3+ + semantic-guided diffusion, and 0.7099 for U-Net + semantic-guided diffusion.

\begin{figure}[!t]
    \centering
    \includegraphics[width=\columnwidth]{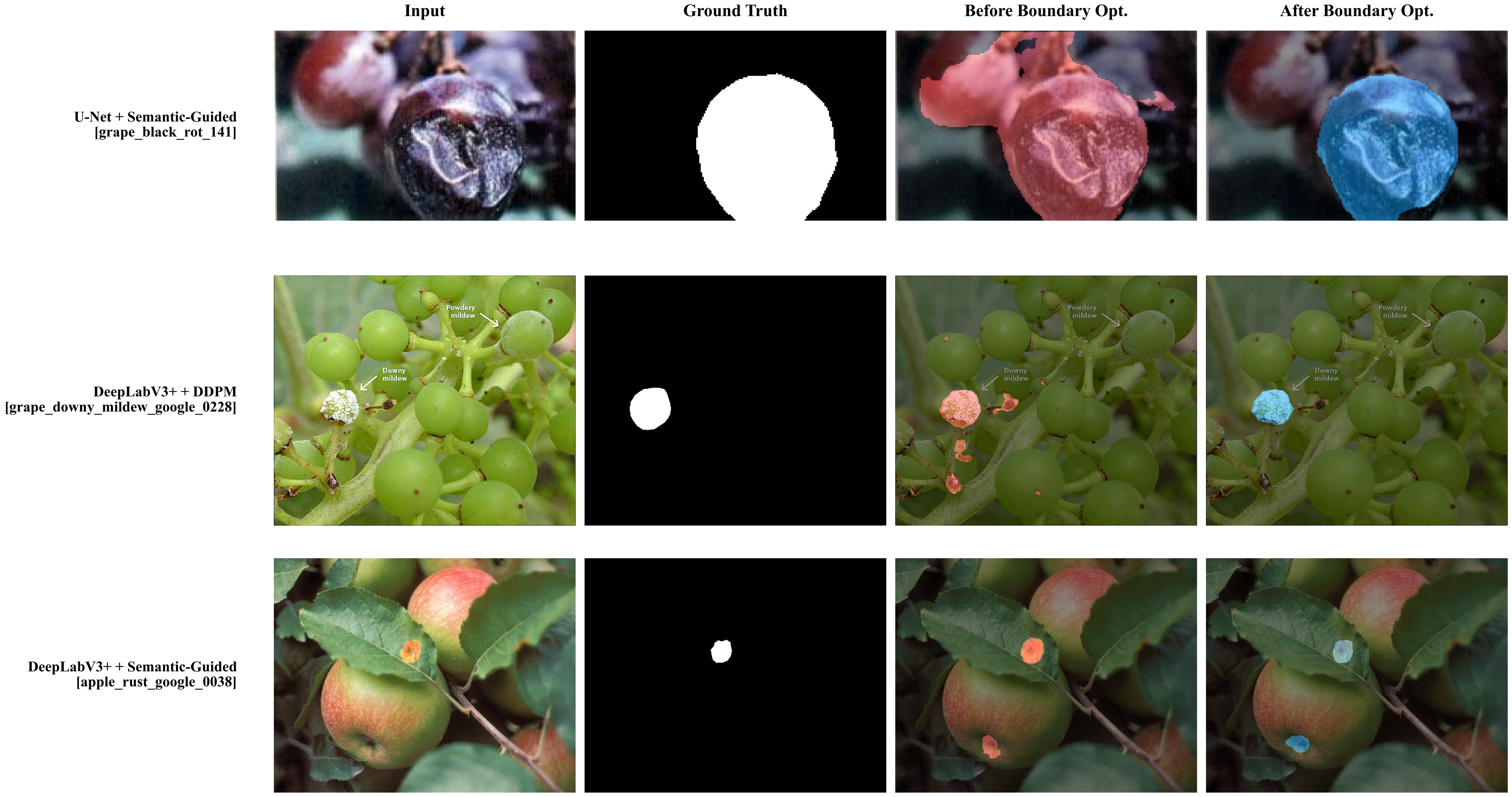}
    \caption{Qualitative comparison of boundary-constrained optimization on representative PlantSegV3 validation samples for the three selected hybrid models. Effect on performance metrics is reported in Table ~\ref{tab:boundary_gain}.
    \label{fig:boundary_constrained}
}
\end{figure}

Table~\ref{tab:boundary_gain} quantifies the effect of boundary-constrained optimization relative to the Stage-1 checkpoints, and Figure~\ref{fig:boundary_constrained} provides the corresponding qualitative examples. The main improvement is not in overlap alone, but in contour fidelity, which is exactly the behavior expected from a boundary-aware objective. DeepLabV3+ + DDPM increases Boundary-F1 from 0.0755 at screening to 0.2194 after boundary-aware optimization. DeepLabV3+ + semantic-guided diffusion rises from 0.0695 to 0.2195, and U-Net + semantic-guided diffusion improves from 0.0797 to 0.2490. These gains are much larger than the changes in mIoU, which indicates that the boundary-aware loss primarily sharpens contour fidelity rather than simply increasing overlap. This behavior is desirable for stress-related plant segmentation, where the target regions are often thin, fragmented, and visually ambiguous \citep{SINGH2018883}.

Among the selected hybrids, DeepLabV3+ + DDPM retains the strongest overlap performance, whereas U-Net + semantic-guided diffusion produces the best boundary localization. DeepLabV3+ + semantic-guided diffusion provides a balanced intermediate behavior. The optimized models therefore support the interpretation that the refiner interacts with the backbone representation in a model-specific way, rather than acting as a generic post-processing block.

\begin{table*}[!t]
\centering
\caption{Effect of boundary-constrained optimization relative to the Stage-1 screening checkpoint. Qualitative comparison is reported in Figure ~\ref{fig:boundary_constrained}.}
\label{tab:boundary_gain}
\resizebox{\textwidth}{!}{%
\begin{tabular}{lcccccc}
\toprule
Model & Screening mIoU & Optimized mIoU & $\Delta$ mIoU & Screening BF1 & Optimized BF1 & $\Delta$ BF1 \\
\midrule
DeepLabV3+ + DDPM & 0.7043 & 0.7114 & +0.0071 & 0.0755 & 0.2194 & +0.1439 \\
DeepLabV3+ + Semantic-Guided & 0.6984 & 0.7052 & +0.0068 & 0.0695 & 0.2195 & +0.1500 \\
U-Net + Semantic-Guided & 0.7026 & 0.7099 & +0.0073 & 0.0797 & 0.2490 & +0.1693 \\
\bottomrule
\end{tabular}%
}
\end{table*}

\FloatBarrier

\subsection{Low-Data Evaluation under Reduced Supervision}
\begin{figure}[!ht]
    \centering
    \includegraphics[width=1\linewidth]{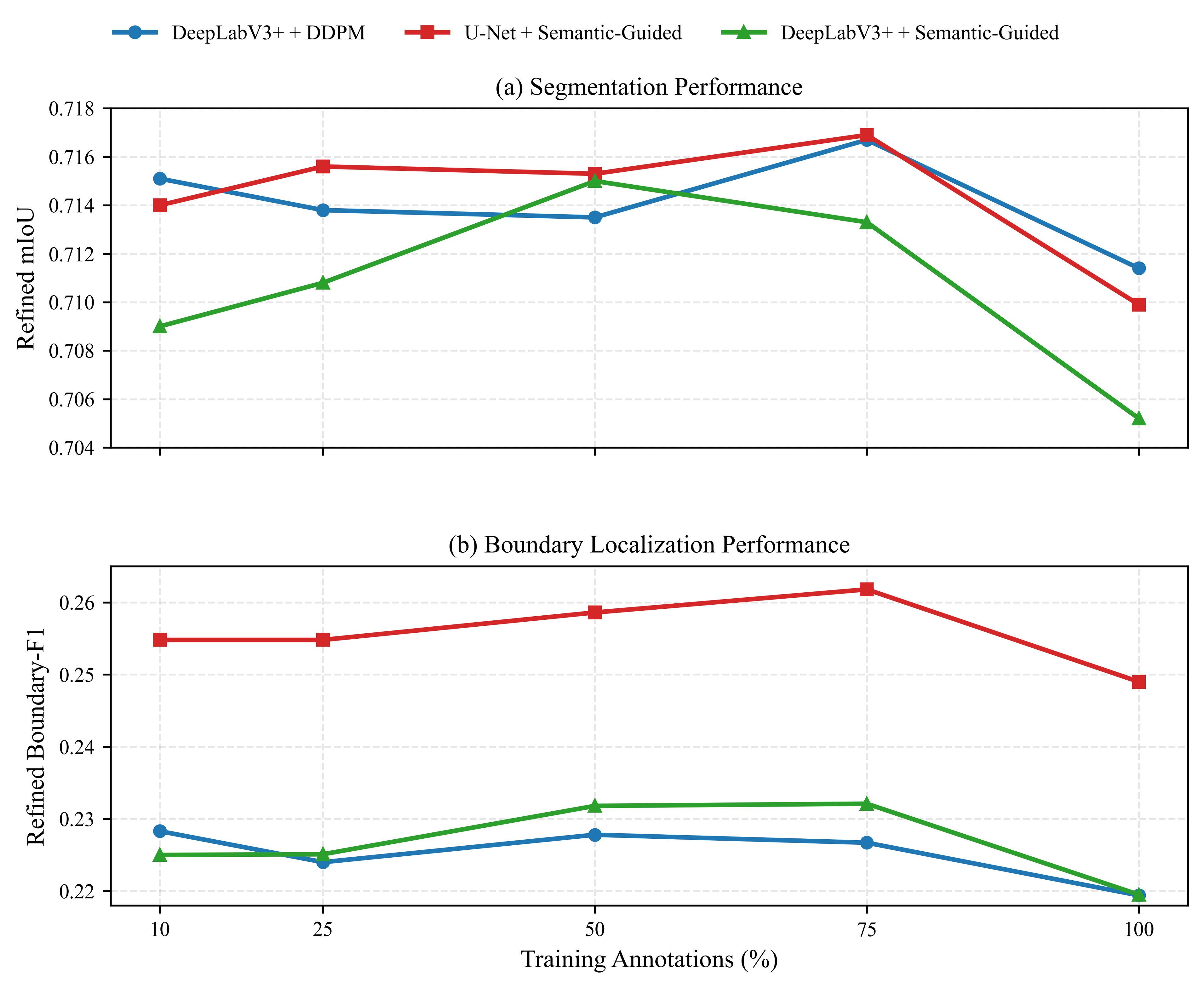}
    \caption{Performance of the selected hybrid segmentation architectures under progressively reduced annotation budgets. (a) Refined mIoU and (b) Refined Boundary-F1 obtained using 100\%, 75\%, 50\%, 25\%, and 10\% of the available training annotations. (See Table~\ref{tab:lowdata}).}
    \label{fig:low_data}
\end{figure}

The selected models are then evaluated under progressively reduced annotation budgets. The results show that performance remains remarkably stable even when supervision is reduced well below the full-data regime. Across the reduced-data settings, the models remain close to the reference condition, and several settings even show small improvements in refined mIoU and Boundary-F1. For example, the observed gains relative to the full-data control remain positive across all three selected hybrids, with the low-data deltas staying in a narrow range rather than collapsing under scarce supervision.
Table~\ref{tab:lowdata} reports the low-data supervision results for the selected hybrid architectures, while Figure~\ref{fig:low_data} visualizes the progression across annotation fractions.
This stability is not due to architecture alone. When the same 10\% supervision setting is repeated with randomly initialized encoders, the performance drops sharply compared with the pretrained counterparts. The pretrained versions achieve refined mIoU values of 0.7151 for DeepLabV3+ + DDPM, 0.7140 for U-Net + semantic-guided diffusion, and 0.7090 for DeepLabV3+ + semantic-guided diffusion, whereas the corresponding randomly initialized models fall to 0.5058, 0.5199, and 0.5097, respectively. The mean gain from pretraining is therefore substantial. This confirms that the strong low-data behavior is primarily driven by the combination of transfer learning and boundary-aware optimization, not by the architectural form alone \citep{he2025masked}.
This result also motivates a study on the appearance diversity of the PlantSegV3 dataset. In the presence of recurring appearance patterns and substantial structural redundancy, even reduced subsets still preserve much of the effective source-domain distribution. In this scenario, the selected hybrids therefore do not require exhaustive annotation to remain useful. Instead, they benefit from a compact but representative supervision set together with pretrained visual priors and a boundary-sensitive objective. This is further explored in subsection 3.8.1.
\begin{table*}[!t]
\centering
\caption{Low-data supervision results for the selected hybrid architectures. The 100\% setting serves as the full-data control, while 75\%, 50\%, 25\%, and 10\% correspond to nested fine-tuning subsets. Best checkpoints are selected using minimum validation loss.}
\label{tab:lowdata}
\small
\renewcommand{\arraystretch}{1.15}
\begin{tabular}{llccccc}
\toprule
Model & Fraction & Samples & Refined mIoU & Refined Dice & Refined BF1 & Pixel Acc \\
\midrule
DeepLabV3+ + DDPM & 100\% & 7916 & 0.7114 & 0.7935 & 0.2194 & 0.9039 \\
DeepLabV3+ + DDPM & 75\%  & 5937 & 0.7167 & 0.7973 & 0.2267 & 0.9072 \\
DeepLabV3+ + DDPM & 50\%  & 3958 & 0.7135 & 0.7944 & 0.2278 & 0.9044 \\
DeepLabV3+ + DDPM & 25\%  & 1979 & 0.7138 & 0.7944 & 0.2240 & 0.9061 \\
DeepLabV3+ + DDPM & 10\%  & 792  & 0.7151 & 0.7963 & 0.2283 & 0.9060 \\
\midrule
U-Net + Semantic-Guided & 100\% & 7916 & 0.7099 & 0.7908 & 0.2490 & 0.9063 \\
U-Net + Semantic-Guided & 75\%  & 5937 & 0.7169 & 0.7971 & 0.2618 & 0.9057 \\
U-Net + Semantic-Guided & 50\%  & 3958 & 0.7153 & 0.7960 & 0.2586 & 0.9044 \\
U-Net + Semantic-Guided & 25\%  & 1979 & 0.7156 & 0.7962 & 0.2548 & 0.9061 \\
U-Net + Semantic-Guided & 10\%  & 792  & 0.7140 & 0.7948 & 0.2548 & 0.9057 \\
\midrule
DeepLabV3+ + Semantic-Guided & 100\% & 7916 & 0.7052 & 0.7860 & 0.2195 & 0.9028 \\
DeepLabV3+ + Semantic-Guided & 75\%  & 5937 & 0.7133 & 0.7931 & 0.2321 & 0.9064 \\
DeepLabV3+ + Semantic-Guided & 50\%  & 3958 & 0.7150 & 0.7955 & 0.2318 & 0.9055 \\
DeepLabV3+ + Semantic-Guided & 25\%  & 1979 & 0.7108 & 0.7918 & 0.2251 & 0.9061 \\
DeepLabV3+ + Semantic-Guided & 10\%  & 792  & 0.7090 & 0.7898 & 0.2250 & 0.9039 \\
\bottomrule
\end{tabular}
\end{table*}

Across all three architectures, performance remains remarkably stable even under substantial reductions in annotation volume, with no evidence of the sharp degradation often expected in low-data regimes. More interestingly, some configurations exhibit slight gains over the full-data reference, indicating that the boundary-constrained objective acts as a beneficial regularizer rather than a mechanism that depends strongly on large-scale supervision. For DeepLabV3+ + DDPM, refined mIoU increases from 0.7114 under full supervision to 0.7151 at 10\%, while Boundary-F1 also rises from 0.2194 to 0.2283. U-Net + Semantic-Guided shows the strongest boundary behavior overall, reaching Boundary-F1 = 0.2586 at 50\% while maintaining refined mIoU above 0.715 across all reduced-data settings. DeepLabV3+ + Semantic-Guided remains similarly robust, with refined mIoU improving to 0.7150 at 50\% and staying close to the full-data control at lower fractions. The strong low-data performance is also supported by the size and visual redundancy of PlantSegV3, which means that even reduced annotation subsets still capture much of the dominant source-domain structure \citep{Li2023Exploiting}.

Overall, these results indicate that the selected hybrid architectures are highly annotation-efficient and that the boundary-aware training regime can maintain, and in some cases slightly improve, both overlap-based and boundary-sensitive performance under limited supervision.

\subsection{Perturbation Sensitivity Analysis}
\begin{figure}[!t]
   \centering
    \includegraphics[width=\columnwidth]{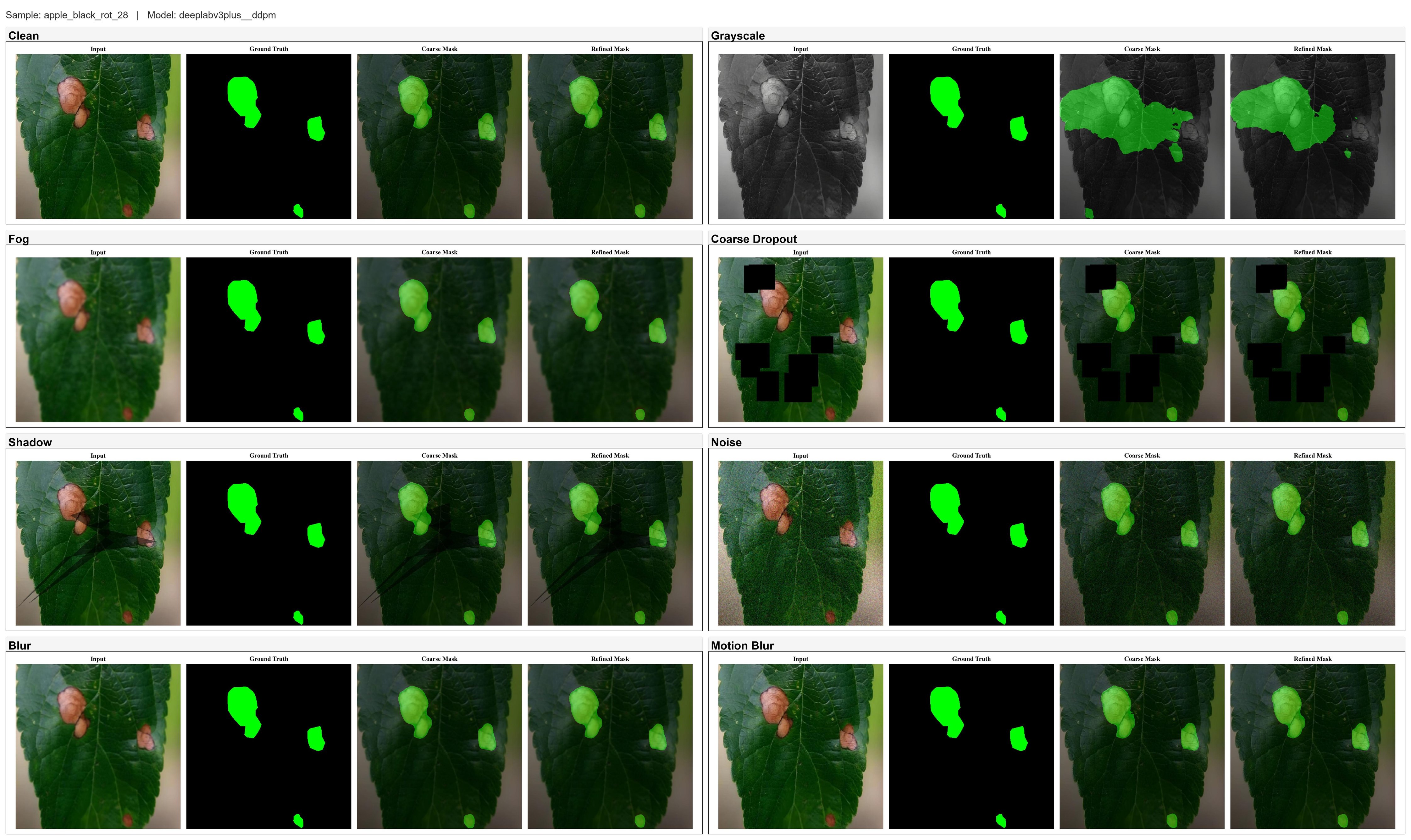}
    \caption{Qualitative robustness assessment of the DeepLabV3+ + DDPM hybrid architecture under representative augmentation-induced perturbations. (See Table~\ref{tab:perturbation_sensitivity_policy}).}
\label{fig:aug} 
\end{figure}

The perturbation analysis identifies the failure modes that most strongly degrade segmentation quality. Among the tested perturbations, grayscale conversion is by far the most damaging, producing a refined mIoU drop of -0.2585. Fog is the second most severe perturbation with a drop of -0.0820, followed by coarse dropout (-0.0590), shadow (-0.0326), Gaussian noise (-0.0244), blur (-0.0209), perspective distortion (-0.0160), brightness-dark shifts (-0.0115), contrast reduction (-0.0100), and motion blur (-0.0082). 
Table~\ref{tab:perturbation_sensitivity_policy} and Figure~\ref{fig:aug} together show that the most damaging perturbations are photometric rather than geometric. Grayscale conversion is the most severe failure mode, followed by fog and coarse dropout, which motivates the augmentation policy used in the retraining stage. The ranking based severity analysis provides a direct empirical basis for the augmentation policy used in the retraining stage.
\begin{sidewaystable*}[p] 
\centering
\caption{
Perturbation sensitivity analysis of the selected DeepLabV3+ + DDPM hybrid model and the augmentation policy derived from the observed performance degradation. 
The table reports refined mIoU and Boundary-F1 under each perturbation, together with the change relative to the clean condition, the resulting severity rank, and the retraining priority assigned for augmentation-guided retraining. 
Negative $\Delta$ values indicate performance loss. Qualitative robustness assessment is reported in Figure ~\ref{fig:aug}.
}
\label{tab:perturbation_sensitivity_policy}
\small
\renewcommand{\arraystretch}{1.15}
\setlength{\tabcolsep}{4pt}
\begin{tabular}{lccccccc}
\toprule
Condition & Refined mIoU & $\Delta$ mIoU & Refined BF1 & $\Delta$ BF1 & Severity rank & Sensitivity level & Retraining priority \\
\midrule
Clean & 0.7066 & 0.0000 & 0.2225 & 0.0000 & -- & -- & -- \\
\midrule
Affine & 0.7111 & +0.0046 & 0.2297 & +0.0072 & -- & Low & Low \\
Horizontal Flip & 0.7060 & -0.0006 & 0.2199 & -0.0027 & -- & Low & Low \\
Elastic & 0.7052 & -0.0014 & 0.2204 & -0.0021 & -- & Low & Low \\
Gamma Bright & 0.7046 & -0.0020 & 0.2182 & -0.0043 & -- & Low & Low \\
Contrast High & 0.7033 & -0.0032 & 0.2199 & -0.0026 & -- & Low & Low \\
Vertical Flip & 0.7018 & -0.0047 & 0.2174 & -0.0051 & -- & Medium & Medium \\
Brightness Bright & 0.7013 & -0.0052 & 0.2162 & -0.0064 & -- & Medium & Medium \\
Gamma Dark & 0.7008 & -0.0058 & 0.2163 & -0.0062 & -- & Medium & Medium \\
Motion Blur & 0.6984 & -0.0082 & 0.2044 & -0.0181 & 10 & Mild & Low \\
Contrast Low & 0.6966 & -0.0100 & 0.2108 & -0.0117 & 9 & Mild & Low \\
Brightness Dark & 0.6951 & -0.0115 & 0.2099 & -0.0126 & 8 & Mild & Low \\
Perspective & 0.6906 & -0.0160 & 0.1948 & -0.0277 & 7 & Moderate & Medium \\
Blur & 0.6857 & -0.0209 & 0.1933 & -0.0293 & 6 & Moderate & Medium \\
Noise & 0.6821 & -0.0244 & 0.1921 & -0.0304 & 5 & Moderate & Medium \\
Shadow & 0.6740 & -0.0326 & 0.1902 & -0.0323 & 4 & Severe & High \\
Coarse Dropout & 0.6476 & -0.0590 & 0.1735 & -0.0490 & 3 & Severe & High \\
Fog & 0.6246 & -0.0820 & 0.1300 & -0.0925 & 2 & Critical & Very High \\
Grayscale & \textbf{0.4481} & \textbf{-0.2585} & \textbf{0.0920} & \textbf{-0.1305} & 1 & Critical & Very High \\
\bottomrule
\end{tabular}
\end{sidewaystable*}

\subsection{Augmentation-Guided Retraining}
\begin{figure}[!t] 
    \centering
    \includegraphics[width=\columnwidth]{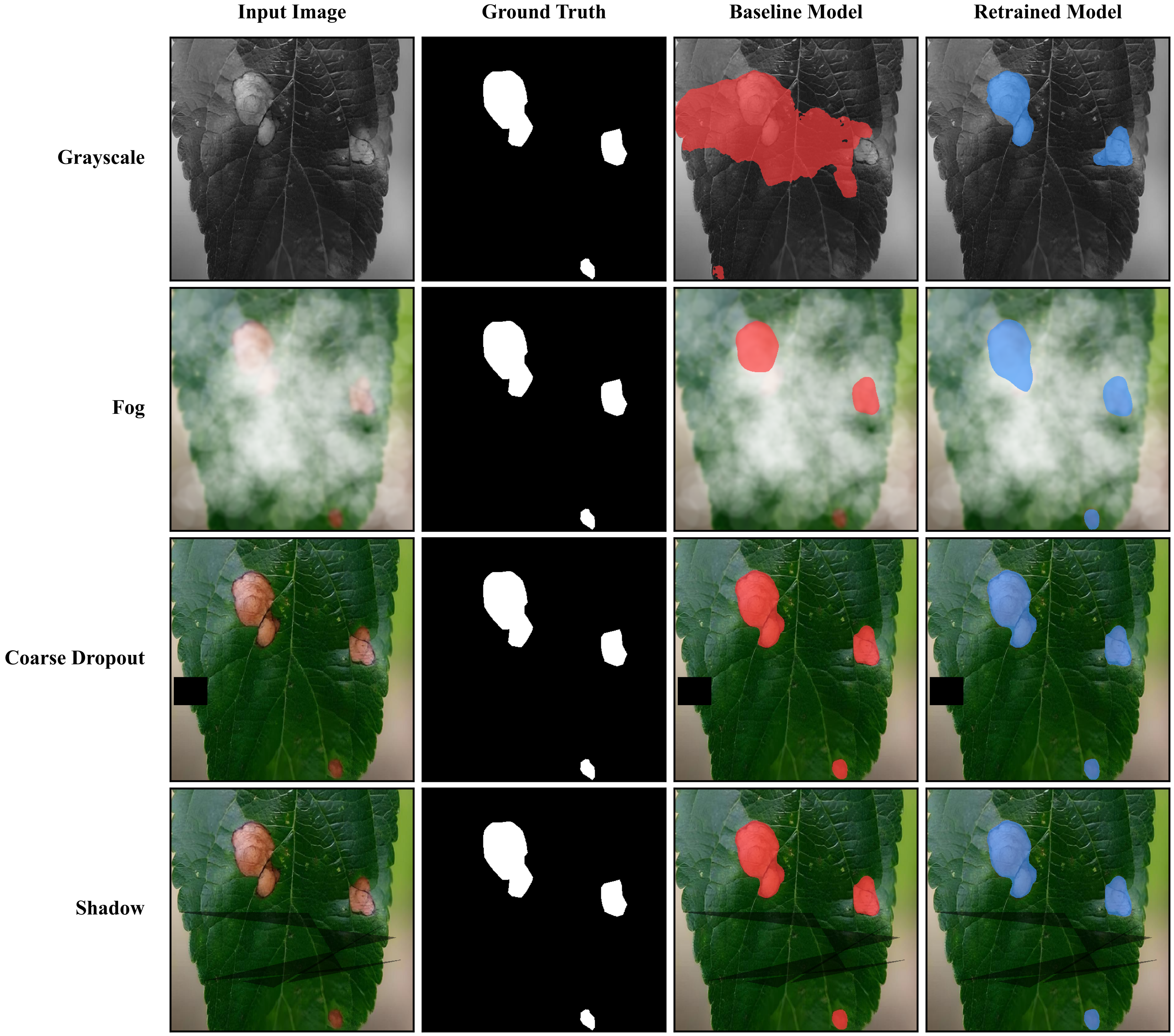}
    \caption{
    Qualitative comparison of the baseline hybrid model and the augmentation-guided retrained model under representative perturbations identified during the perturbation sensitivity analysis. Quantitative improvement in performance after retraining is reported in Table ~\ref{tab:ddpm_before_after_aug}.}
    \label{fig:augmentation_guided_retraining_qualitative}
\end{figure}

The perturbation-derived augmentation policy is used to retrain the selected models on PlantSegV3. The retraining stage improves robustness while preserving the clean-domain performance of the optimized models. On the clean validation split, the retrained checkpoints achieve refined mIoU values of 0.7165 for DeepLabV3+ + DDPM, 0.7122 for DeepLabV3+ + semantic-guided diffusion, and 0.7111 for U-Net + semantic-guided diffusion. Boundary-F1 also remains strong, with values of 0.2352, 0.2273, and 0.2601, respectively.
The augmentation policy improves robustness most clearly on the perturbations that were previously identified as severe. Fog and shadow are recovered to near-clean levels, while coarse dropout and noise remain more challenging but still noticeably improved compared with the unadapted models. Grayscale remains the hardest condition, which is consistent with the sensitivity ranking and confirms that the models still depend partly on color information.
Table~\ref{tab:ddpm_before_after_aug} summarizes the robustness gains obtained after augmentation-guided retraining, and Figure~\ref{fig:augmentation_guided_retraining_qualitative} shows representative qualitative recovery under the most severe perturbations. The retraining policy improves robustness without degrading clean-domain performance, indicating that the perturbation-derived augmentation curriculum acts as a corrective mechanism rather than a generic regularizer.

\begin{table*}[!t]
\centering
\caption{
Effect of augmentation-guided retraining on the robustness of the DeepLabV3+ + DDPM hybrid architecture. Positive $\Delta$ values indicate performance improvement after retraining. Qualitative comparison is reported in Figure ~\ref{fig:augmentation_guided_retraining_qualitative}.
}
\label{tab:ddpm_before_after_aug}
\small
\renewcommand{\arraystretch}{1.12}
\begin{tabular}{lcccccc}
\toprule
Condition &
Before mIoU &
After mIoU &
$\Delta$ mIoU &
Before BF1 &
After BF1 &
$\Delta$ BF1 \\
\midrule

Clean             & 0.7066 & 0.7165 & +0.0099 & 0.2225 & 0.2352 & +0.0127 \\
Grayscale         & 0.4481 & 0.6375 & +0.1894 & 0.0920 & 0.1550 & +0.0630 \\
Fog               & 0.6246 & 0.7153 & +0.0907 & 0.1300 & 0.2342 & +0.1042 \\
Coarse Dropout    & 0.6476 & 0.6832 & +0.0356 & 0.1735 & 0.1985 & +0.0250 \\
Shadow            & 0.6740 & 0.7096 & +0.0356 & 0.1902 & 0.2281 & +0.0379 \\
Noise             & 0.6821 & 0.6988 & +0.0167 & 0.1921 & 0.2098 & +0.0177 \\
Blur              & 0.6857 & 0.6997 & +0.0140 & 0.1933 & 0.2118 & +0.0185 \\
Motion Blur       & 0.6984 & 0.7098 & +0.0114 & 0.2044 & 0.2208 & +0.0164 \\
Brightness Dark   & 0.6951 & 0.7059 & +0.0108 & 0.2099 & 0.2252 & +0.0153 \\
Brightness Bright & 0.7013 & 0.7083 & +0.0070 & 0.2162 & 0.2277 & +0.0115 \\
Contrast Low      & 0.6966 & 0.7125 & +0.0159 & 0.2108 & 0.2284 & +0.0176 \\
Contrast High     & 0.7033 & 0.7087 & +0.0054 & 0.2199 & 0.2291 & +0.0092 \\
Gamma Dark        & 0.7008 & 0.7099 & +0.0091 & 0.2163 & 0.2275 & +0.0112 \\
Gamma Bright       & 0.7046 & 0.7148 & +0.0102 & 0.2182 & 0.2342 & +0.0160 \\
Affine            & 0.7111 & 0.7192 & +0.0081 & 0.2297 & 0.2435 & +0.0138 \\
Perspective       & 0.6906 & 0.7000 & +0.0094 & 0.1948 & 0.2060 & +0.0112 \\
Elastic           & 0.7052 & 0.7149 & +0.0097 & 0.2204 & 0.2346 & +0.0142 \\
\bottomrule
\end{tabular}
\end{table*}

\subsection{Constrained Hyperparameter Screening}
A short hyperparameter search is then performed around the validated retraining regime. The goal is not to pursue exhaustive optimization, but to check whether the retrained models can be fine-tuned further under a restricted compute budget \citep{RAIAAN2024100470}. The results show only modest improvements, indicating that augmentation-guided retraining already places the models close to a stable operating point. 

After constrained screening, U-Net + semantic-guided diffusion reaches the best clean-validation mIoU of 0.7183 and Boundary-F1 of 0.2658. DeepLabV3+ + semantic-guided diffusion improves to 0.7164 mIoU and 0.2447 BF1, while DeepLabV3+ + DDPM remains essentially unchanged in mIoU at 0.7161 but improves in BF1 to 0.2410. These changes are best interpreted as fine calibration rather than as a shift in performance regime. In other words, the augmentation-guided retraining stage has already captured the main benefit, and the short hyperparameter search merely adjusts the operating point. This result is useful for the manuscript because it supports the resource-aware design philosophy. The pipeline does not depend on brute-force search or extensive compute. Instead, it reaches a stable solution through a short screening process, targeted retraining, and a narrow hyperparameter pass. The best performing hyperparameter configurations for all three selected hybrid architectures are reported in Table ~\ref{tab:hpo_search_space_best}.
\begin{table*}[!t]
\centering
\caption{The best configuration selected for each hybrid architecture after constrained hyperparameter screening. The search is intentionally narrow and centered around the validated augmentation-guided retraining regime to keep the optimization budget low while preserving methodological consistency.}
\label{tab:hpo_search_space_best}
\renewcommand{\arraystretch}{1.12}
\setlength{\tabcolsep}{6pt}
\small
\begin{tabular}{@{}lccc@{}}
\toprule
Parameter & DeepLabV3+ & DeepLabV3+ & U-Net \\
& + DDPM & + Semantic-Guided & + Semantic-Guided \\
\midrule
Learning rate & $3\times10^{-5}$ & $5\times10^{-5}$ & $3\times10^{-5}$ \\
Weight decay & $10^{-4}$ & $10^{-4}$ & $0$ \\
Batch size & $8$ & $8$ & $4$ \\
Dice weight & $1.25$ & $1.0$ & $1.5$ \\
Boundary weight & $1.25$ & $1.0$ & $1.25$ \\
Lovász weight & $0.5$ & $1.0$ & $1.0$ \\
Auxiliary loss weight & $0.2$ & $0.2$ & $0.2$ \\
Gradient clipping & $1$ & $1$ & $3$ \\
\bottomrule
\end{tabular}
\end{table*}

\subsection{Cross-Domain Transfer and Controlled Adaptation}

To evaluate the transferability of the learned representations, the augmentation-guided models were adapted to two external benchmarks with progressively increasing target supervision. Unlike the source-domain experiments, these evaluations intentionally expose the models to substantial distribution shifts, allowing us to assess not only direct transfer but also the recoverability of performance under controlled adaptation.

Direct transfer produces a pronounced reduction in segmentation performance on both datasets, confirming that the source-domain representations do not generalize unchanged across agricultural domains. On NWRD, the best zero-shot checkpoint achieves 0.5186 refined mIoU and 0.0482 Boundary-F1, while on CWFID the corresponding scores fall further to 0.1090 and 0.0257, respectively. Although both datasets differ from PlantSegV3, the nature of the domain shift is fundamentally different. NWRD remains a disease-segmentation benchmark with similar lesion localization objectives, whereas CWFID introduces structured crop–weed field imagery together with a different annotation ontology. Consequently, transfer to CWFID requires adaptation not only to new appearance statistics but also to new semantic classes and scene organization, making it a substantially more challenging benchmark.

Introducing even a small amount of target supervision progressively restores segmentation quality. On NWRD, the refined masks improve steadily as the support fraction increases, eventually reaching 0.5892 mIoU and 0.2195 Boundary-F1 after full adaptation. The qualitative examples demonstrate that the transferred models already preserve coarse lesion localization during zero-shot inference, while adaptation mainly improves contour precision and suppresses false-positive responses.

\begin{figure}[!t]
\centering
\includegraphics[width=\columnwidth]{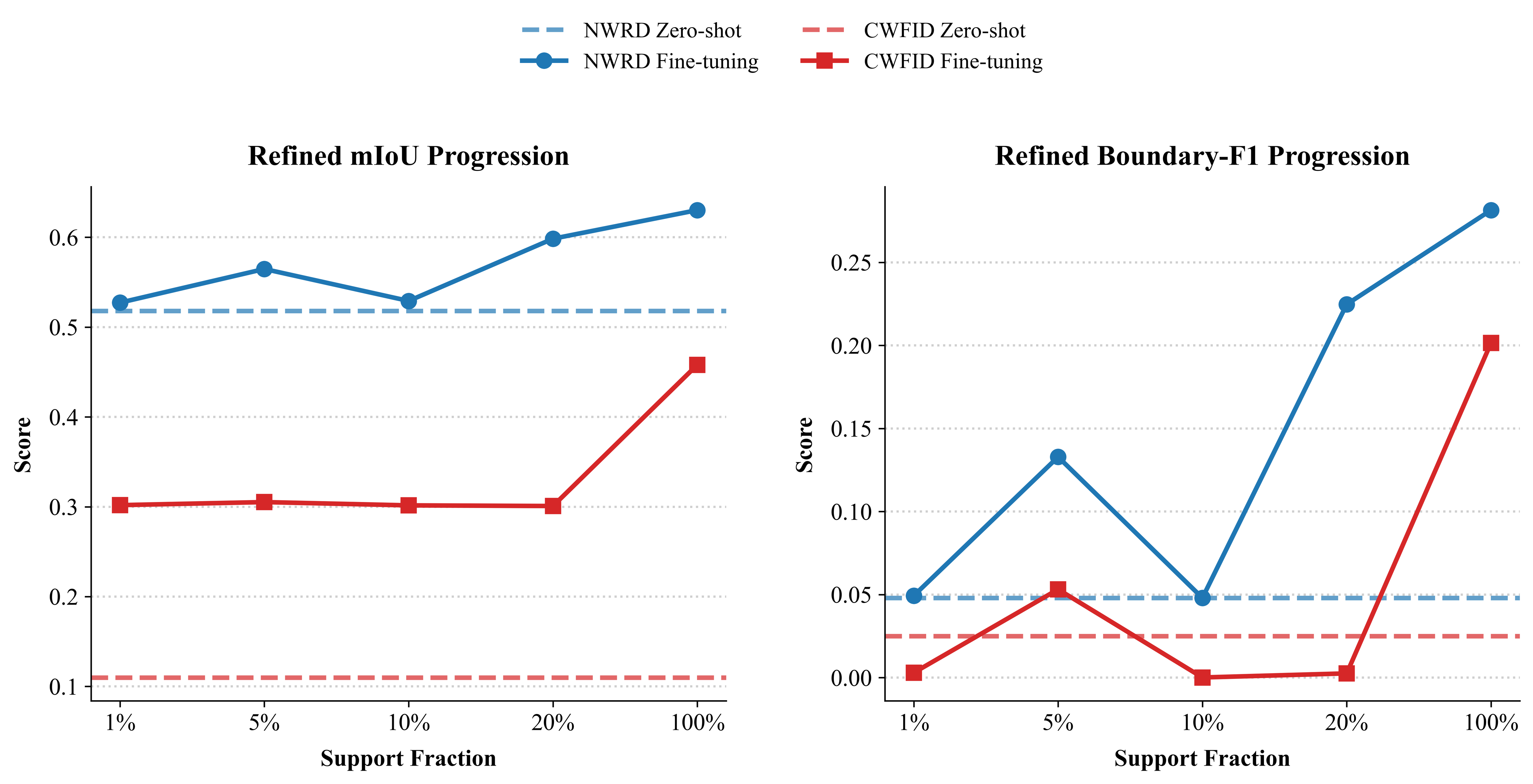}
\caption{
    Effect of increasing target-domain supervision on cross-domain adaptation performance. The best-performing hybrid architecture under each adaptation regime was evaluated using progressively larger target-domain support sets (1\%, 5\%, 10\%, 20\%, and full supervision).}
    \label{fig:fewshot_curves}
\end{figure}

\begin{figure}[!t]
    \centering
    \includegraphics[width=\columnwidth]{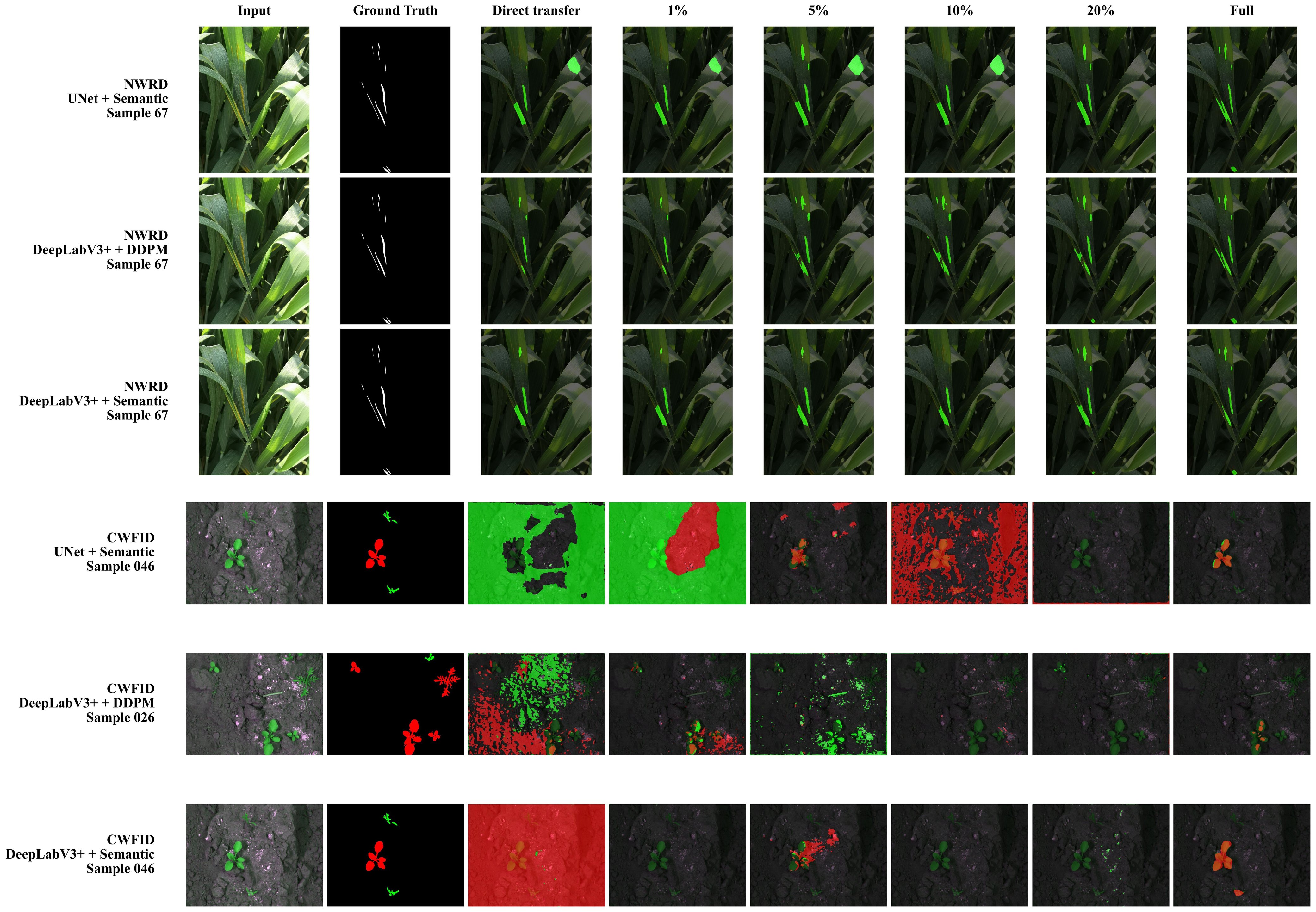}
    \caption{
    Qualitative evaluation of controlled few-shot adaptation under cross-domain transfer. For each target dataset, predictions are shown before adaptation (direct transfer) and after adaptation using progressively larger fractions of labeled target-domain data (1\%, 5\%, 10\%, 20\%, and full supervision). Best performing checkpoints are reported in Table~\ref{tab:zero_shot_vs_adapted}.}
\label{fig:fewshot_adaptation_qualitative}
\end{figure}

A different adaptation behavior is observed on CWFID. Zero-shot predictions largely fail to distinguish crop and weed structures because the source models were optimized for stress-region localization rather than semantic crop/weed discrimination. Nevertheless, increasing target supervision enables the models to progressively reorganize their learned feature representations. The qualitative results show a gradual transition from fragmented foreground activation during zero-shot inference to increasingly coherent crop and weed segmentation after adaptation, illustrating that the pretrained representations retain transferable structural information despite the substantial task mismatch.
Table~\ref{tab:zero_shot_vs_adapted} reports the representative zero-shot and adapted checkpoints, Figure~\ref{fig:fewshot_curves} shows how performance evolves as target supervision increases, and Figure~\ref{fig:fewshot_adaptation_qualitative} provides the corresponding qualitative recovery patterns.

Importantly, full-shot adaptation does not completely eliminate the performance gap relative to the source-domain results. This behavior is expected because the models are initialized from representations learned predominantly on PlantSegV3 and subsequently fine-tuned on the target datasets rather than trained from scratch. Consequently, the final representations remain partially influenced by the source-domain optimization trajectory, particularly for CWFID where both visual statistics and annotation semantics differ substantially from the source task. These findings indicate that the proposed hybrid models learn transferable structural priors that can be efficiently adapted to new domains, while also highlighting the influence of source-domain bias under large appearance and task shifts.

\begin{table*}[!t]
\centering
\caption{Representative best-performing checkpoints before and after controlled target adaptation. The zero-shot column reports the best direct-transfer checkpoint from the augmentation-guided retrained models, while the adapted column reports the best full fine-tuning result for each benchmark. Effect of increasing target-domain supervision can be observed in Figure ~\ref{fig:fewshot_curves}, while qualitative performance reported in Figure ~\ref{fig:fewshot_adaptation_qualitative}.}
\label{tab:zero_shot_vs_adapted}
\renewcommand{\arraystretch}{1.15}
\setlength{\tabcolsep}{5pt}
\small
\begin{tabular}{llllcc}
\toprule
Dataset & Regime & Best model & Refined mIoU & Refined BF1\\
\midrule
NWRD & Zero-shot & DeepLabV3+ + Semantic-Guided & 0.5186 & 0.0482\\
NWRD & Full fine-tuning & DeepLabV3+ + DDPM & 0.5892 & 0.2195\\
CWFID & Zero-shot & U-Net + Semantic-Guided & 0.1090 & 0.0257\\
CWFID & Full fine-tuning & U-Net + Semantic-Guided & 0.4573 & 0.2216\\
\bottomrule
\end{tabular}
\end{table*}

\subsection{Discussion}
\begin{figure}[!t]
    \centering
    \includegraphics[width=\columnwidth]{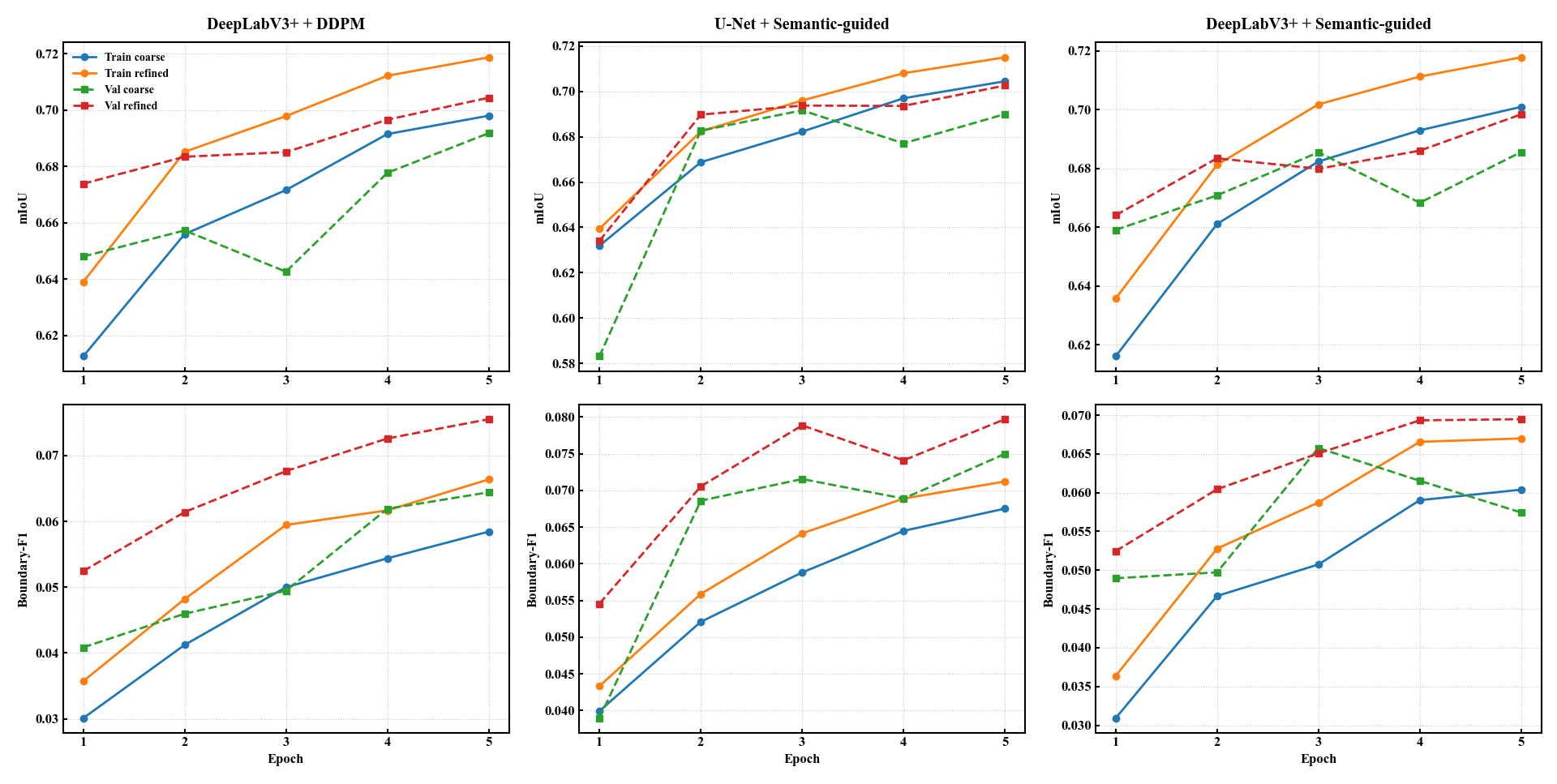}
    \caption{
Training dynamics of the selected hybrid configurations during the Stage-1 screening process. Curves are shown for both coarse and diffusion-refined predictions, illustrating convergence behavior and the evolution of validation mIoU and Boundary-F1 across epochs.}
    \label{fig:training_dynamics}
\end{figure}

Taken together, the results support four main conclusions. First, diffusion refinement is beneficial when used as a structural correction stage for coarse segmentation masks, as evidenced by the consistent performance margins maintained between coarse and refined predictions throughout the training process (Figure ~\ref{fig:training_dynamics}). Second, boundary-aware optimization produces much larger gains in contour fidelity than in overlap alone, which is important for plant stress structures that are thin, fragmented, or partially occluded. Third, low-data robustness is preserved because pretrained backbones and the boundary-sensitive objective provide strong inductive bias, allowing the model to remain stable even with limited supervision. Fourth, direct cross-domain transfer fails without adaptation, but controlled target supervision restores a substantial fraction of the lost performance.

The results also show why the resource-constrained design matters. The best-performing configurations were identified through short screening runs on a local RTX 4050 6 GB GPU rather than through exhaustive training. The paper demonstrates that a careful combination of architectural screening, perturbation-aware retraining, and targeted adaptation can yield robust segmentation behavior without requiring large-scale compute \citep{10.3389/fpls.2023.1308528}.

From a modeling perspective, the three refinement strategies behave differently. DDPM is strongest when the priority is overlap stability \citep{Hoetal2020}, semantic-guided diffusion is most effective when boundary localization matters \citep{Liuetal2023}, and latent diffusion is less reliable under the current short-screening and limited-budget regime \citep{Rombachetal2022}. This suggests that diffusion priors are useful, but only when the conditioning mechanism is aligned with the segmentation backbone and the training regime. The framework therefore provides not just a performance comparison, but a practical selection rule for compute-aware plant segmentation.

\subsubsection{Dataset Redundancy and Appearance Diversity}

\begin{figure}[!t]
    \centering
    \includegraphics[width=\columnwidth]{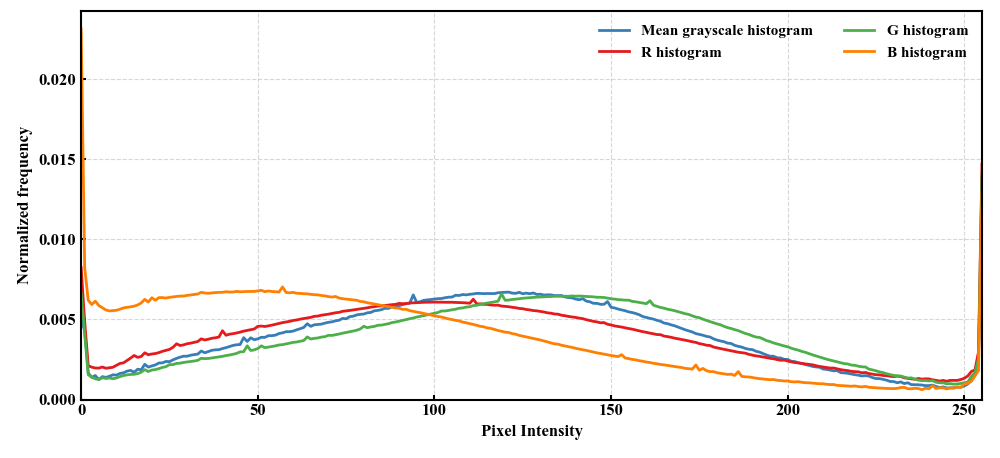}
    \caption{Mean grayscale and RGB histogram profiles computed across the PlantSegV3 training split. The distributions exhibit a dominant concentration within the mid-intensity range while retaining non-negligible variability across the full dynamic range.
}
\label{fig:mean_hist}
\end{figure}

\begin{figure}[!ht]
    \centering
    \includegraphics[width=\linewidth]{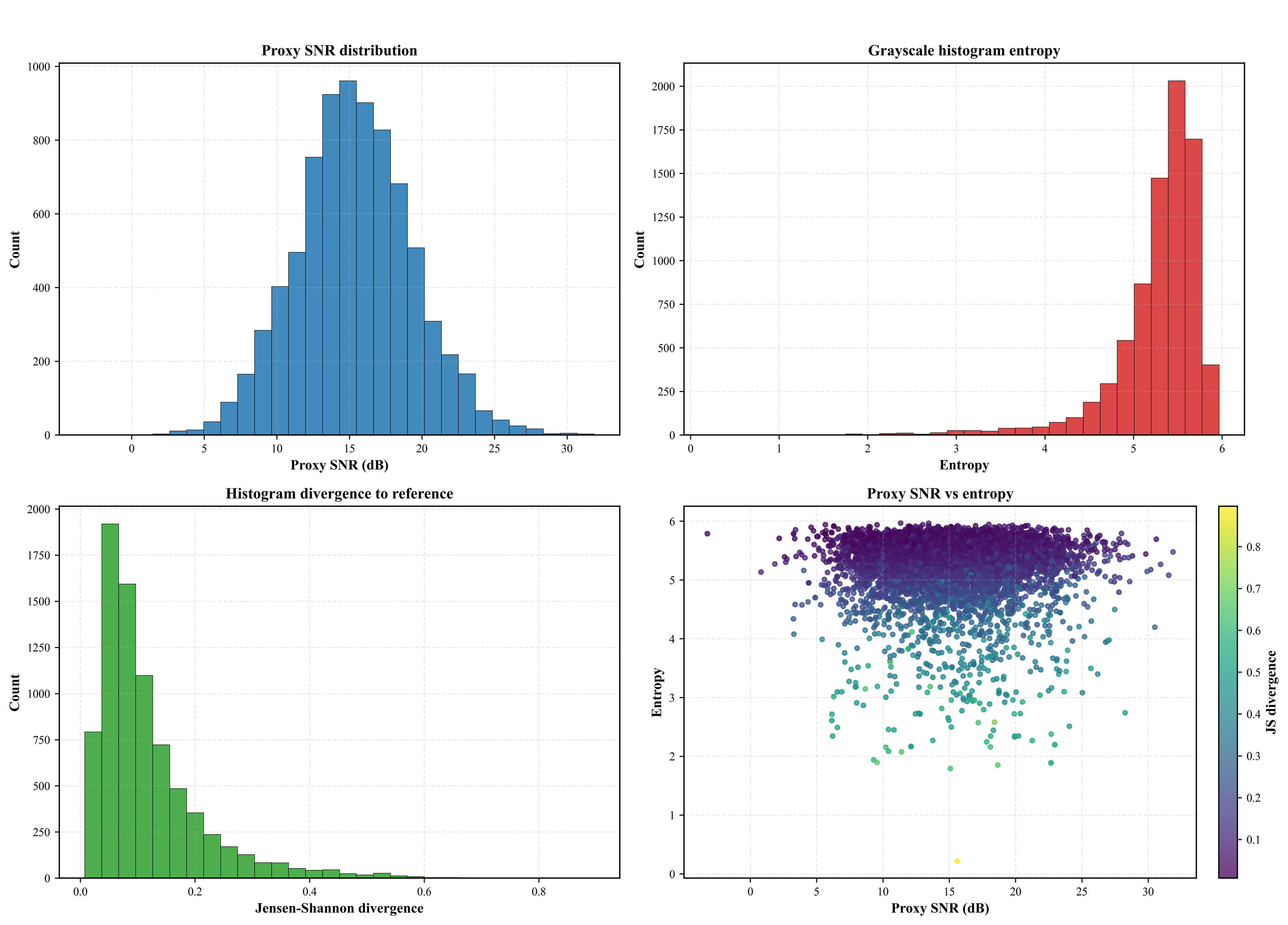}
    \caption{Redundancy and appearance-diversity analysis of the PlantSegV3 training split. Top-left: distribution of proxy signal-to-noise ratio (SNR). Top-right: grayscale histogram entropy distribution. Bottom-left: Jensen-Shannon divergence of individual image histograms relative to the dataset reference histogram. Bottom-right: joint distribution of proxy SNR and entropy, coloured by histogram divergence.
}
\label{fig:dash}
\end{figure}

\begin{figure}[!t]
    \centering
    \includegraphics[width=\columnwidth]{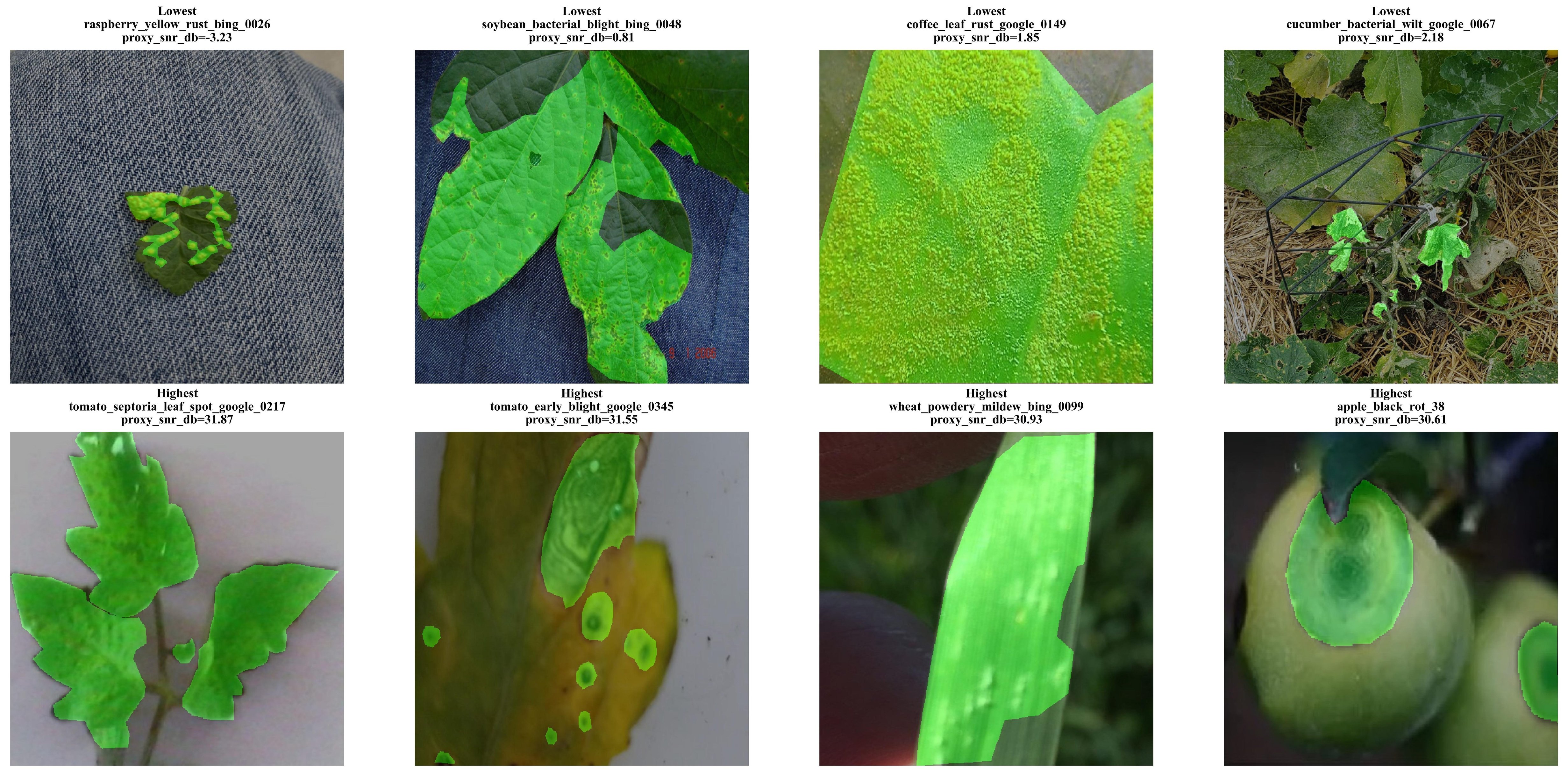}
    \caption{Representative training samples selecrted from the lower and upper extremes of the proxy SNR distribution. The low-SNR examples exhibit reduced contrast, increased background clutter, and more challenging acquisition conditions, whereas the high-SNR examples show clearer object boundaries and improved visual separability.
}
\label{fig:panel_comp}
\end{figure}

The relatively strong low-data performance observed in this study motivated an additional appearance-level analysis of PlantSegV3. The goal of this analysis was not to suggest that the dataset is visually simple, but to determine whether it contains recurring intensity patterns, texture statistics, and morphological structures that make the learning problem partially redundant. Such redundancy is important in low-data settings because a reduced annotation subset can still represent much of the dominant visual distribution if the dataset is internally repetitive \citep{Hussain2026Quality,Li2023Exploiting}.
Figure~\ref{fig:mean_hist} shows the mean grayscale and Red, Green, Blue (RGB) histogram profiles. The normalized distributions exhibit a dominant mass in the mid-intensity range together with visible extreme-intensity tails, suggesting a mix of recurrent plant appearance and illumination-driven variation. Figure~\ref{fig:dash} complements this view by combining proxy signal-to-noise ratio, histogram entropy, and Jensen-Shannon divergence \citep{zhang2024jscdscoredataselection,10.1145/2505515.2505756,app112110088}, while Figure~\ref{fig:panel_comp} shows examples from the low and high SNR extremes. Taken together, these figures indicate that PlantSegV3 is heterogeneous but not uniformly dispersed in appearance space, which helps explain why the selected models remain stable under reduced supervision.
\\\\We first computed normalized grayscale and RGB histogram profiles over the training split. The broad overlap among samples indicates that many images share similar global intensity statistics, while the sharp peaks near the low and high intensity extremes reflect the coexistence of large dark background regions, shadowed areas, and saturated foreground pixels \citep{HOLUB2006620}. In other words, the dataset contains a strong dominant appearance mode, together with extreme-intensity tails that are typical of field imagery and illumination variation. This combination suggests that the model can learn a stable core representation from a limited number of annotations, while still encountering nontrivial appearance variation.
\\\\To quantify image-level variation more explicitly, we computed the Jensen-Shannon (JS) divergence between each sample histogram and the reference histogram of the full training set. Lower JS divergence indicates that a sample closely follows the common dataset distribution, whereas higher values identify more atypical appearance patterns \citep{Rahnenfuhrer2023Statistical,zhang2024jscdscoredataselection}. In parallel, a proxy signal-to-noise ratio (proxy SNR) was used to assess how clearly the foreground structure stands out from the surrounding background clutter, and histogram entropy was used as a measure of appearance spread \citep{snr}.The joint distribution of proxy SNR and entropy shows that most samples occupy a compact region of moderate entropy and moderate separability, which is consistent with the presence of recurring image patterns and moderate redundancy. At the same time, a smaller set of low-SNR samples exhibits heavier clutter, weaker contrast, and less favorable acquisition conditions, confirming that the dataset is diverse enough to remain challenging.
\\\\Representative samples from the lower and upper ends of the proxy SNR distribution further support this interpretation. Low-SNR examples typically contain more background clutter, weaker object contrast, and less distinct boundaries, whereas high-SNR examples exhibit cleaner structural separation and more easily identifiable foreground regions \citep{app112110088}. Importantly, both groups still preserve similar plant morphology and texture motifs, indicating that PlantSegV3 combines global redundancy with local variability. This explains why the selected hybrid models remain reasonably stable under reduced supervision: even when the annotation budget is reduced, the remaining samples still cover much of the dominant visual structure of the dataset.

\subsubsection{Effect of Encoder Pretraining}
The low-data results in Table~\ref{tab:lowdata} show that the selected hybrid models remain robust even at 10\% supervision, but this behavior is not explained by architecture alone. To isolate the contribution of transfer learning, the 10\% setting was repeated with randomly initialized encoders while keeping the same objective and training schedule fixed \citep{Stuckner2022Microstructure,DINI2026115850}. Table~\ref{tab:pretraining_ablation} shows that the pretrained models substantially outperform the random-initialization counterparts, with a mean gain of 0.2009 mIoU. This confirms that the strong low-data behavior is driven by the combination of pretrained visual priors and boundary-aware optimization rather than by the model topology alone.
\begin{table*}[!t]
\centering
\caption{
Effect of encoder pretraining under extreme annotation scarcity (10\% supervision). The pretrained models correspond to the low-data supervision experiments using ImageNet-pretrained encoders, while the random-initialization models are trained from scratch under the same annotation fraction.
}
\label{tab:pretraining_ablation}
\renewcommand{\arraystretch}{1.15}
\setlength{\tabcolsep}{5pt}
\begin{tabular}{lccc}
\toprule
Model &
Pretrained mIoU &
Random mIoU &
$\Delta$ mIoU \\
\midrule
DeepLabV3+ + DDPM &
0.7151 &
0.5058 &
+0.2093 \\

U-Net + Semantic-Guided &
0.7140 &
0.5199 &
+0.1941 \\

DeepLabV3+ + Semantic-Guided &
0.7090 &
0.5097 &
+0.1993 \\
\midrule
\textbf{Mean Gain} &
-- &
-- &
\textbf{+0.2009} \\
\bottomrule
\end{tabular}
\end{table*}

\subsection{Contextual Benchmarking}

Table~\ref{tab:benchmark_context} provides a contextual comparison between representative benchmark studies and the best checkpoints obtained in this work. The comparison is useful for positioning the proposed framework relative to prior results, but it should not be interpreted as a direct quantitative comparison because the datasets, splits, label spaces, and evaluation protocols are not identical across studies. On PlantSegV3, the proposed hybrid model improves over the published reference results under the same source-domain benchmark \citep{Weietal2026}, supporting the benefit of diffusion-guided refinement and boundary-aware optimization. 

For the transfer benchmarks, the table should be read as a measure of recoverability rather than direct in-domain comparability. On NWRD \citep{s23156942}, the adapted checkpoint reflects target-domain recovery under a different acquisition context and patch-based protocol. On CWFID, the lower absolute scores are expected because the task involves a substantial annotation-paradigm shift, where the supervision changes from stress-phenotype segmentation to crop-weed scene understanding \citep{haug15}. In this sense, the table helps interpret the external-transfer results as evidence of transferable structural priors under domain and label-space shift, rather than as a direct comparison with the original benchmark settings.

\begin{table*}[t]
\centering
\caption{Contextual benchmark comparison between representative published studies and the best checkpoints from the proposed framework. Because the datasets, splits, label spaces, and evaluation protocols differ across studies, the numbers are not intended for strict cross-paper ranking.}
\label{tab:benchmark_context}
\small
\renewcommand{\arraystretch}{1.2}
\begin{tabularx}{\textwidth}{L{1.8cm} L{1.8cm} L{3.5cm} L{3.5cm} Y}
\toprule
Benchmark dataset & Reference study & Performance reported & Proposed diffusion-hybrid framework & Evaluation protocol \\
\midrule

PlantSegV3 & \citet{Weietal2026} &
ConvNeXt-L: 46.24\% mIoU, 59.97\% mAcc; SegNext MSCAN-L: 44.52\% mIoU, 59.95\% mAcc. &
U-Net + Semantic-Guided: 71.83\% refined mIoU, 26.10\% refined BF1. &
Source-domain benchmark under constrained hyperparameter screening; the resulting checkpoint after augmentation-guided retraining serves as the source model for all downstream transfer experiments. \\
\addlinespace

NWRD & \citep{s23156942} &
Rust class: F1 = 0.638, precision = 0.621, recall = 0.675. &
DeepLabV3+ + DDPM: 58.92\% refined mIoU, 21.95\% refined BF1. &
Controlled few-shot adaptation on an external disease-segmentation benchmark; the result reflects target-domain recovery rather than direct in-domain training. \\
\addlinespace

CWFID & \citet{AGUSTIAN2026100357} &
Teacher: 94.98\% mIoU / 97.40\% F1; distilled student: 86.95\% mIoU / 98.79\% pixel accuracy. &
U-Net + Semantic-Guided: 45.73\% refined mIoU, 22.16\% refined BF1. &
Few-shot adaptation under a substantial annotation-paradigm shift, where the semantic label space changes from stress-phenotype segmentation to crop--weed scene understanding. \\

\bottomrule
\end{tabularx}
\end{table*}

\subsubsection{Limitations and Future Work}

This work has several limitations. First, the study is intentionally focused on source-domain screening and controlled adaptation rather than on exhaustive optimization over a large architecture or hyperparameter search space. This design choice was necessary to keep the experiments feasible under local compute constraints, but it also means that the reported results should be interpreted as resource-aware findings rather than fully saturated upper bounds \citep{RAIAAN2024100470,10.3389/fpls.2023.1308528,resource.aware}. Second, the present study evaluates stress-related segmentation at the level of region delineation and robustness, but it does not yet extend to downstream structural phenotyping descriptors or fine-grained biological trait analysis. Third, although the selected models transfer to a reasonable extent after adaptation, direct zero-shot transfer remains challenging ~\ref{tab:zero_shot_vs_adapted}, which confirms that stronger domain invariance is still needed.

Future work will therefore move beyond source-domain robustness and focus on the downstream utilization of the recovered segmentation masks for more detailed structural phenotyping and biological trait analysis \citep{horticulturae12050640,LIU2026100164}. In particular, we will investigate whether diffusion-refined masks can support richer phenotypic characterization and whether alternative refinement and transfer strategies can further improve generalization across diverse plant domains. An important direction will be the integration of knowledge distillation to transfer the structural knowledge learned by the proposed hybrid models into student networks. \citet{AGUSTIAN2026100357} demonstrated the viability of such architectures by successfully distilling a heavy DeepLabV3+ network into a lightweight Fast Spatial Convolutional Neural Network (SCNN) student for efficient plant-related semantic segmentation enabling efficient deployment without significant loss of segmentation quality. In addition, future studies will explore more structure-preserving segmentation strategies, adaptive domain transfer mechanisms, and the biological consistency of the recovered masks under increasingly challenging domain shifts.

\section{Conclusion}

This study investigated whether diffusion-guided refinement can improve stress-related plant segmentation under realistic constraints on training budget, supervision, and domain shift. Using PlantSegV3 as the source benchmark, we evaluated a $3 \times 3$ space of backbone-refiner combinations and found that the effectiveness of diffusion refinement depends strongly on the underlying segmentation backbone and conditioning strategy. Among the screened configurations, DeepLabV3+ + DDPM, U-Net + semantic-guided diffusion, and DeepLabV3+ + semantic-guided diffusion emerged as the most stable and useful hybrid models. These selected architectures consistently improved upon their coarse predictions, and boundary-constrained optimization further strengthened structural delineation without degrading region-level overlap.

A key outcome of this work is that competitive segmentation performance can be achieved under a compact training budget. The models were screened and retrained on a local RTX 4050 6 GB GPU using short training schedules rather than exhaustive optimization, yet the resulting pipeline remained effective across full-data, low-data, perturbation-aware, and cross-domain settings. Low-data evaluation showed that the selected models remain stable under reduced supervision, while perturbation-guided augmentation retraining improved resilience to realistic appearance shifts. Direct transfer to external datasets caused substantial performance degradation, but controlled target-domain adaptation recovered much of the lost overlap and boundary quality. These findings indicate that diffusion is effective as a structural refinement mechanism when paired with an appropriate backbone and trained under a robustness-oriented protocol.

Overall, the proposed framework offers a resource-aware route to plant segmentation that is not limited to a single benchmark setting. Rather than optimizing only for in-domain accuracy, the study emphasizes structural reliability, label efficiency, and transferability under constrained computation. This makes the framework relevant for agricultural segmentation problems where annotated data are limited and operating conditions vary across domains.
\section*{CRediT Authorship Contribution Statement}
\textbf{Gurbhit Chaurakoti:} Conceptualization, Methodology, Software, Data curation, Visualization, Validation, Writing -- original draft, Writing -- review \& editing. \textbf{Soumyashree Kar:} Conceptualization, Methodology, Supervision, Investigation, Validation, Visualization, Formal analysis, Writing -- review \& editing, Project administration.

\section*{Declaration of Competing Interest}
The authors declare that they have no known competing financial interests or personal relationships that could have appeared to influence the work reported in this paper.

\section*{Data Availability}

The data used in this study are derived from publicly available, open-source repositories. Brief descriptions of each dataset are provided below.

\textbf{PlantSegV3 Dataset:} The PlantSegV3 dataset provides in-the-wild plant disease segmentation masks, containing 7,774 diseased images across 34 host plants and 69 disease types \citep{Weietal2026}. It is publicly available and can be accessed at \url{https://github.com/tqwei05/PlantSeg}.

\textbf{NUST Wheat Rust Disease (NWRD) Dataset:} The NWRD dataset serves as an external transfer benchmark for disease-related localization and segmentation, containing 100 high-resolution wheat-rust images under a different acquisition context \citep{s23156942}. It is publicly available at \url{https://github.com/dll-ncai/NUST-Wheat-Rust-Disease-NWRD}.

\textbf{Crop/Weed Field Image Dataset (CWFID):} The CWFID dataset introduces structured crop-weed scenes, containing 60 top-down field images with 162 crop plants and 332 weed plants annotated with vegetation masks and crop/weed labels \citep{haug15}. This dataset is publicly available at \url{https://github.com/cwfid/dataset}.

\section*{Code Availability}
The code will be available upon reasonable request from the corresponding author.
\bibliographystyle{elsarticle-harv}
\bibliography{ref}

\vfill
\end{document}